\documentclass[10pt,journal,compsoc]{IEEEtran}
\newenvironment{packed_item}{
\begin{itemize}
  \setlength{\itemsep}{1pt}
  \setlength{\parskip}{2pt}
  \setlength{\parsep}{0pt}
}{\end{itemize}}

\usepackage{bbm}

\ifCLASSOPTIONcompsoc
  \usepackage[nocompress]{cite}
\else
  \usepackage{cite}
\fi
\ifCLASSINFOpdf
  \usepackage[pdftex]{graphicx}
\else
\fi
\usepackage{amsmath}
\begin{document}
%
% paper title
% Titles are generally capitalized except for words such as a, an, and, as,
% at, but, by, for, in, nor, of, on, or, the, to and up, which are usually
% not capitalized unless they are the first or last word of the title.
% Linebreaks \\ can be used within to get better formatting as desired.
% Do not put math or special symbols in the title.
\title{Learning Intrinsic Images for Clothing}
%
%
% author names and IEEE memberships
% note positions of commas and nonbreaking spaces ( ~ ) LaTeX will not break
% a structure at a ~ so this keeps an author's name from being broken across
% two lines.
% use \thanks{} to gain access to the first footnote area
% a separate \thanks must be used for each paragraph as LaTeX2e's \thanks
% was not built to handle multiple paragraphs
%
%
%\IEEEcompsocitemizethanks is a special \thanks that produces the bulleted
% lists the Computer Society journals use for "first footnote" author
% affiliations. Use \IEEEcompsocthanksitem which works much like \item
% for each affiliation group. When not in compsoc mode,
% \IEEEcompsocitemizethanks becomes like \thanks and
% \IEEEcompsocthanksitem becomes a line break with idention. This
% facilitates dual compilation, although admittedly the differences in the
% desired content of \author between the different types of papers makes a
% one-size-fits-all approach a daunting prospect. For instance, compsoc 
% journal papers have the author affiliations above the "Manuscript
% received ..."  text while in non-compsoc journals this is reversed. Sigh.

\author{Kuo Jiang,
        Zian Wang,
        Xiaodong Yang% <-this % stops a space
}

\IEEEtitleabstractindextext{%
\begin{abstract}
Reconstruction of human clothing is an important task and often relies on intrinsic image decomposition. 
% Clothing images are challenging cases for intrinsic image decomposition models, usually containing detailed shape deformation and complex texture. 
With a lack of domain-specific data and coarse evaluation metrics, existing models failed to produce satisfying results for graphics applications. 
% and the evaluation metrics are also too coarse to provide thorough and diagnostic information on edges. 
In this paper, we focus on intrinsic image decomposition for clothing images and have comprehensive improvements. 
We collected CloIntrinsics, a clothing intrinsic image dataset, including a synthetic training set and a real-world testing set. 
A more interpretable edge-aware metric and an annotation scheme is designed for the testing set, which allows diagnostic evaluation for intrinsic models. 
Finally, we propose ClothInNet model with carefully designed loss terms and an adversarial module. 
It utilizes easy-to-acquire labels to learn from real-world shading, significantly improves performance with only minor additional annotation effort. 
We show that our proposed model significantly reduce texture-copying artifacts while retaining surprisingly tiny details, outperforming existing state-of-the-art methods. 
\end{abstract}

% Note that keywords are not normally used for peerreview papers.
}

% make the title area
\maketitle

% To allow for easy dual compilation without having to reenter the
% abstract/keywords data, the \IEEEtitleabstractindextext text will
% not be used in maketitle, but will appear (i.e., to be "transported")
% here as \IEEEdisplaynontitleabstractindextext when the compsoc 
% or transmag modes are not selected <OR> if conference mode is selected 
% - because all conference papers position the abstract like regular
% papers do.
\IEEEdisplaynontitleabstractindextext
% \IEEEdisplaynontitleabstractindextext has no effect when using
% compsoc or transmag under a non-conference mode.

% For peer review papers, you can put extra information on the cover
% page as needed:
% \ifCLASSOPTIONpeerreview
% \begin{center} \bfseries EDICS Category: 3-BBND \end{center}
% \fi
%
% For peerreview papers, this IEEEtran command inserts a page break and
% creates the second title. It will be ignored for other modes.
\IEEEpeerreviewmaketitle

\IEEEraisesectionheading{\section{Introduction}\label{sec:intro}}

% % TODO!!!
% \begin{figure}[t!]
% \centering
% \includegraphics[width=0.9\linewidth]{src/teaser.pdf} 
% \caption{\footnotesize\textbf{Clothing intrinsic image decomposition and image editing.} 
% On these challenging cases, we perform color style editing and colorize the clothing with a single color. Our ClothInNet model significantly reduces texture-copying artifacts and thus allows realisitic editing. 
% }
% \label{fig:teaser}
% \end{figure}

Analysis and reconstruction of human clothing texture is an important task for computer vision and computer graphics, yet still receiving little specialized attention. 
In recent works, intensive research efforts have been focus on analysis of human appearance, e.g., reconstruction of 3D human~\cite{imber2014intrinsic,wu2011shading,guo2017real} and editing or synthesis of human characters \cite{meka2016live,li2013capturing}. Many of these methods leverage intrinsic image decomposition \cite{imber2014intrinsic} and related techniques like shape from shading \cite{wu2011high} to decompose texture information into reflectance, geometry and lighting information. 
As a fact that clothing covers most of the surface area of people in daily life, it is vital to focus on clothing texture.
% However, all these methods regard the human texture as a whole, and do not exclusively focus on analysis of human clothing appearance. 

In this paper, we consider the problem of intrinsic image decomposition for human clothing. 
% Intrinsic image decomposition was extensively studied in the past 40 years, 
Driven by the real-world applications like image editing and benefits for other vision tasks such as 3D reconstruction and image segmentation, intrinsic image decomposition was extensively studied but remains a challenging task due to its ill-posed nature. 
It aims to decompose a given image $I$ into corresponding intrinsic images, usually the reflectance image $R$ encoding the surface albedo and the shading image $S$ encoding illumination conditions, satisfying Hadamard product $I = R \odot S$. 

Early work designed energy terms based on priors and formulated the task as an energy minimization problem \cite{land1971lightness,omer2004color,bousseau2009user,rother2011recovering,garces2012intrinsic,zhao2012closed,liao2013non}. 
However, they usually suffer from large computational cost, and the proposed priors are not always true due to the complex conditions of wild images. 
An attractive alternative is to exploit the powerful CNNs and learn the decomposition in a data-driven manner \cite{narihira2015direct,shi2017learning,nestmeyer2017reflectance,baslamisli2018cnn,fan2018revisiting,cheng2018intrinsic,li2018cgintrinsics}. 
However, different from other vision tasks, intrinsic images are extremely hard to annotate. 
% This is because human can only provide relative comparison instead of dense annotation. 
Existing largescale datasets are either synthetic data with dense groundtruth \cite{butler2012naturalistic,li2018cgintrinsics,baslamisli2018joint,baslamisli2018cnn} or real world images with sparse annotation \cite{bell2014intrinsic,kovacs2017shading}. 
With the data-hungry bottleneck, efficient usage of diverse data sources \cite{janner2017self,li2018learning,ma2018single} for indirect supervision becomes more important. 

Recent state-of-the-art methods with higher benchmark scores still suffer from severe texture-copying artifacts, stopping them from real world applications. 
The bottleneck mainly lies in \emph{metric} and \emph{data}. 
With research progress in this field, existing popular evaluation metrics \cite{bell2014intrinsic,kovacs2017shading,grosse2009ground} can easily get cheated and no longer truly reflect real performance \cite{nestmeyer2017reflectance}. Models with better quantitative scores may perform qualitatively worse, making it urgent to have a more expressive metric. 
For the data-hungry bottleneck, to gain enough supervision for this ill-posed task, it's essential to intelligently utilize diverse data sources.% to further improve current models' performance. 
% Also, the evaluation metrics for intrinsic images seldom include edge performance and are not interpretable. 
% With the coarse evaluation, a model with high scores can still produce fairly noisy results. 
% fail on challenging cases like clothing images, which usually contain rather detailed shape and complex texture. The predicted intrinsic images suffers from severe texture-copying artifacts, stopping them from real world applications. 

In this paper, we address the weaknesses mentioned above and aim to improve the performance on the challenging clothing images. 
% of intrinsic image decomposition and have comprehensive 
% in \emph{data}, \emph{metric} and \emph{model}. 
First, to relieve the data-hungry situation of clothing intrinsic images, we collected CloIntrinsics dataset, with a rendered synthetic clothing image training set and a testing set with real world clothing images. Professional software for clothing design \cite{Cloth3D} is used to simulate as realistic as possible clothing models. 
% which are then rendered with diverse texture, different viewpoints and random illumination conditions. 
Second, we carefully designed an annotation scheme and propose a new edge-aware metric to evaluate intrinsic image decomposition models. It targets at common artifacts on real world images and provide interpretable and diagnostic feedback. 
% An annotation scheme is carefully designed to give much clearer indications on common artifacts. 
% The testing set, with a carefully designed annotation scheme, allows to evaluate model performance on both regions and edges. The diagnostic metrics are able 
Third, we propose ClothInNet, a simple but effective model for clothing intrinsic image decomposition. 
In this model, we incorporate the large number of easy-to-acquire \emph{single-color clothing} images as additional supervision sources. 
These images reflect the probability distrubution of real-world shading data, which is learned by the network in an adversarial manner. 
% and utilize an adversarial learning module to enforce the output shading to follow the distribution of real-world shading. 
Additionally, a novel loss term is proposed to apply local gradient constraint on reflectance images and significantly improves performance. 
% It carefully utilized the priors of clothes, and significantly improves the performance with only minor additional annotation. 

The contribution of this paper can be summarized as: 
\begin{packed_item}
	\item a clothing intrinsic image dataset, including a synthetic training set and a real world testing set (Sec.~\ref{sec:dataset}),
	\item a diagnostic metric to evaluate intrinsic image decomposition models (Sec.~\ref{sec:metric}), 
	\item a model to distil real-world shading distribution and carefully designed loss terms to apply exclusive gradient constraints on albedo (Sec.~\ref{sec:method}).
	 % from easy-to-acquire labels and a simple but effective CNN model utilizing these data  
	% proposed a weakly supervised model based on adversarial learning, which achieves better shading performance on clothing images. 
\end{packed_item}

\section{Related Work}

% The research on intrinsic image originates from the study of human vision in the 1970s. Visual system is a crucial part of human perception of the external environment. The visual effects of objects observed by human beings are affected by many factors, such as material, texture, shape, position and color of light source, and position of the observer. 
% From experimental results, scientists observed that the human visual system has the ability to separate light and color. 
% To explain this phenomenon, Land \etal \cite{land1971lightness} proposed Retinex theory on human color perception. 
% Barrow \etal \cite{barrow1978recovering} then formally proposed the task of intrinsic image decomposition.
 % which decomposes a color image into different attributes and represents it as a series of intrinsic images. 
\vspace*{-1mm}
\textbf{Intrinsic Image Decomposition Methods.}
Intrinsic image decomposition algorithms can be roughly divided into two categories: (1) optimization algorithms based on hand-crafted priors and (2) data-driven learning-based methods. 
Early methods usually formulate this task as an optimization problem \cite{land1971lightness,omer2004color,shen2008intrinsic,bousseau2009user,rother2011recovering,garces2012intrinsic,zhao2012closed,liao2013non,barron2014shape,liu2020unsupervised,zhu2020learning,li2021sparse}. 
The main idea is to design a specific energy function with prior knowledge observed from image data, and then solve for the intrinsic images by minimizing the energy function. 
But real-world images can not fully meet these hand-crafted priors, and thus severe artifacts would occur. 
% Also, optimization-based methods generally suffer from high computational cost, making them less appealing for applications in real-time. 
% In recent years, a large number of data-driven learning-based methods, especially deep learning methods, have emerged in the field of computer vision and computer graphics. 
% Researchers also proposed learning-based methods for intrinsic image decomposition. 
With fruitful achievements in other vision tasks, a promising alternative is to learn data-driven features, leveraging the powerful CNNs for feature extraction \cite{narihira2015direct,shi2017learning,nestmeyer2017reflectance,janner2017self,li2018learning,ma2018single,fan2018revisiting,cheng2018intrinsic,lettry2018darn,sengupta2019neural,yu2019inverserendernet}. 
These works generally consumes RGB images as input and uses a CNN to predict one or both intrinsic images. 
Proposed methods like mirror-link architecture \cite{shi2017learning}, post-processing \cite{nestmeyer2017reflectance}, intermediate representations \cite{janner2017self} and adversarial module \cite{lettry2018darn} are shown to improve the results. 
% DirectIntrinsics\cite{narihira2015direct} proposed the Multiscale CNN Regression method and directly predict reflectance and shading from RGB. 
% Shi \etal\cite{shi2017learning} employed a mirror-link CNN architecture to predict intrinsic images and expanded to non-Lambertian settings. 
% Nestmeyer \etal\cite{nestmeyer2017reflectance} designed a CNN fully trained with sparse annotation, and found that traditional filtering methods with strong prior could still significantly increase the performance of CNN outputs. 
To handle the scarcity in the amount and domain of annotated data, recent research shifts the focus to 
% generalization ability and seek for 
indirect sources of supervision, including self-supervision \cite{janner2017self}, shared modules across datasets \cite{fan2018revisiting}, differentiable renderer \cite{sengupta2019neural,yu2019inverserendernet}, multiview images \cite{duchene2015multi,ma2018single} and multi-illumination sequences depicting the same scene \cite{lettry2018unsupervised,li2018learning,bi2018deep}. 
% \cite{janner2017self} explored using self-supervision for generalization and \cite{fan2018revisiting} proposed a network with modules that can largely shared across datasets. 
% \cite{lettry2018unsupervised,li2018learning,bi2018deep} employed the image sequences depicting the same scene but under different illumination as training data, and thus learn without ground truth decomposition. 
% \cite{duchene2015multi,ma2018single} leveraged multiview images of the scene for 3D geometry recovery to better understanding the intrinsic and lighting information.  
% \cite{sengupta2019neural,yu2019inverserendernet} explores differentiable renderer and use self-supervision to train intrinsic image decomposition models. 
However, current state-of-the-art models still generates severe artifacts for richly textured clothing images. 
The major problem is that we can't densely annotate real-world images, thus lacking the knowledge of real-world intrinsic images. 
Our proposed method utilizes clothing priors to learn real-world shading distribution and proposes a novel gradient constraint loss to address this. 
% Our ClothInNet model, with an adversarial module, can utilize easy-to-annotate labels to efficiently acquire real-world shading images. Different from \cite{lettry2018darn} in design, the adversarial module is employed to learn real-world shading distribution. 

\noindent\textbf{Intrinsic Images Datasets.}
Unlike other vision tasks, it's difficult to obtain largescale densely annotated intrinsic images, as people can't directly annotate the absolute value of reflectance and shading. 
% Also, many people may never ``see'' the reflection and shading images in their life, and this also bring great challenges to accurate data collection. 
Existing largescale datasets lie in two categories: (1) real-world images with sparse annotation and (2) synthetic data with dense groundtruth. 
% A more practical compromise is to annotate the relative comparision of pixel pairs. 
Intrinsic Images in the Wild (IIW) dataset \cite{bell2014intrinsic} provides sparse annotation for the pairwise comparison of reflectance on real world scene images. 
Shading Annotations in the Wild (SAW) dataset \cite{kovacs2017shading} provides additional sparse shading annotation on the same image data. 
However, due to the sparse labeling, the data itself is difficult to train a high-performance CNN model \cite{narihira2015direct,nestmeyer2017reflectance}. 
Synthetic datasets such as MPI Sintel \cite{butler2012naturalistic}, ShapeNet \cite{chang2015shapenet} and CGIntrinsics  \cite{li2018cgintrinsics} are also popular for training CNNs with dense supervision. 
% \cite{baslamisli2018joint} also proposed to jointly learn intrinsic images and semantic segmentation. 
But synthetic data essentially suffer from a domain gap and will affect the performance on real world data. 
% But people can easily tell apart real world images from CG rendered data, indicating a gap between their distribution. 
% The bottleneck for current learning based methods is the lack of data. 
As far as we know, CloIntrinsics dataset we propose in this work is the first intrinsic image dataset in clothing domain. 
With diverse shape deformation, lighting and texture, CloIntrinsics improves current models to produce tightly fitted predictions. 

\noindent\textbf{Intrinsic Image Metrics.}
In addition to qualitative comparison, quantitative metrics on real world datasets facilitate evaluation and comparison of intrinsic image decomposition algorithms. 
MIT Intrinisic Images dataset \cite{grosse2009ground} provides dense groundtruth intrinsic images for 20 objects, which enables quantitative comparison of existing algorithms. 
% Considering the scale ambiguity of intrinsic images, 
It uses scale-invariant mean square error as the quantitative metric. % for intrinsic algorithms. 
For real-world scene images, IIW dataset \cite{bell2014intrinsic} employs weighted human disagreement rate as a metric for reflectance, which tells how well the predicted results agrees with the sparse human annotation. 
SAW dataset \cite{kovacs2017shading} classifies the shading into smooth and non-smooth, and uses precision-recall curve to evaluate the results. 
These existing metrics are coarse and lack interpretable evaluation on edge performance. We contribute an annotation scheme and metric to address these problems.

\section{CloIntrinsics Dataset} 
\label{sec:dataset}
% \INFO{
% Sketch: Introduce the datasets. 

% Intuition as intro: why datasets? 

% subsec1: synthetic training set

% subsec2: testing set and metric definition 
% }

\begin{figure*}
    \centering
    \includegraphics[width=\linewidth]{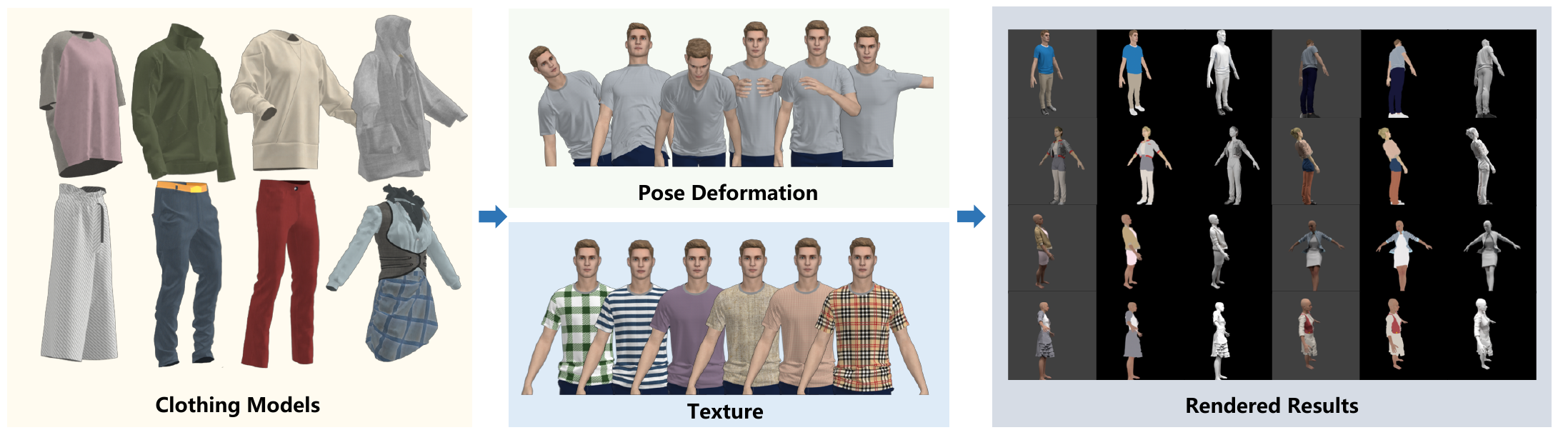} 
    \caption{Visualization of CloIntrinsics Training Set. We collect 3D clothing models designed by artists and put them on avatars. 
    The pose of avatars is changed to simulate various wrinkle patterns in real world, and the texture patterns are changed to generate diverse reflectance. 
    These models are rendered with different viewpoints and lighting and form the synthetic training set. }
    \label{fig:CloIntrinsics}
\end{figure*}

% Since data is the bottleneck for intrinsic models, 
% As far as we know, there's no intrinsic image dataset in clothing domain. 
To relieve the data-hungry situation and improve the relatively rough evaluation metrics for intrinsic models, we propose Clothing Intrinsic Images (CloIntrinsics) Dataset, including a synthetic training set and a real-world testing set. 
% We carefully design the annotation scheme and propose a diagnostic metric for intrinsic models. 
\subsection{Synthetic Training Set}
Although largescale synthetic datasets for animated film \cite{butler2012naturalistic} and indoor scenes \cite{li2018cgintrinsics} exist, we believe clothes generally have different priors in the shape deformation and texture patterns, and the in-domain data will surely improve the performance of intrinsic image decomposition models. 
The key is to render as realistic as possible clothing data. 
We choose to use fashion design software CLO3d \cite{Cloth3D}, which produce realistic 3D clothing models with garment flats, and render the 3D models with Blender \cite{Blender}. 
To simulate real world conditions better, as shown in Fig.~\ref{fig:CloIntrinsics}, the training set contains diversity in the following aspects. 

\noindent\textbf{Shape.} 
% Real-world clothing images can have complex and detailed shapes, and recovering the correct shading effect is crucial in many applications. 
The variety of clothing shapes generally comes from different clothing types, and the shape deformation causes by the wearer's poses. 
To cover common clothing types like T-shirt, jacket, dress, suit, etc., CloIntrinsics collects 80 3D clothing models designed by artists. 
To include diverse wrinkle patterns, we change the poses of avatars and simulate the deformation of clothing. 
In this way, it collects diverse shape patterns for clothing. 
As a further augmentation, we render these models in different viewpoints to get more diverse images. 

\noindent\textbf{Reflectance.} 
The complex texture is among the challenges for clothing intrinsic image decomposition and it's vital to render a dataset with diverse texture. 
To address this, we build a database with around 200 texture patterns. 
As fashion design software like CLO3d designs clothing with garment flats, we change the texture pattern of garment flats for each 3d clothing models, and thus increase diversity in reflectance. 

\noindent\textbf{Lighting.}
Real world images generally contain complex lighting conditions. 
In addition to global ambient lights, we generate random light sources during rendering to improve the complexity of lighting conditions. 
The light sources consist of one hemisphere lighting and 10 to 20 point lights randomly sampled on the upper hemisphere, with random intensity. 

% \vspace{1mm}
The data generation pipeline is shown in Fig.\ref{fig:CloIntrinsics}. We first collect 3D clothing models designed by artists. We then put them on avatars and simulate the shape deformation caused by pose changes. Different texture patterns are used to include diverse reflectance map. We finally render the models with different viewpoints and lighting conditions. 
With the settings above, we finally raytraced 5K clothing images with paired intrinsic components at the resolution of 1024x1024. 
% These data contain prior knowledge in clothing domain and improve the performance of intrinsic image decomposition models. 

\subsection{Real World Testing Set} 
To convincingly reflect the real-world performance, we choose to evaluate our model on real world clothing images instead of synthetic data. 
We randomly select 500 clothing images from Paper Doll Dataset \cite{yamaguchi2013paper}, and ask human experts with knowledge background in intrinsic images to annotate \emph{regions} and \emph{edges} of interest. 
The proposed edge-aware metric significantly improves the interpretability and expressiveness of existing metrics. 
% To improve the interpretability and expressive of existing metrics, we propose a novel edge-aware metric and design an corresponding annotation scheme. 
The details are included in Sec.~\ref{sec:metric} below.

\section{Edge-Aware Metric}
\label{sec:metric}

% Generally speaking, a good evaluation metric should reflect how models perform against common artifacts. 
The task of intrinsic image decomposition requires to distinguish whether the local pixel value changes, i.e. edges, are caused by the change of reflectance or shading. 
Also, the edge performance of intrinsic image decomposition models is crucial in many graphics applications, and deserves special attention. 
However, current popular evaluation metrics are not able to evaluate edge performance in an interpretable way, and thus not expressive to show how models perform against artifacts. 
For instance, the major challenge for intrinsic models is the texture-copying artifact, which often happens on edges due to the entanglement of intrinsic components. The sparse WHDR metric in IIW dataset \cite{bell2014intrinsic} and AP used in SAW dataset \cite{kovacs2017shading} are not able to reflect this artifact. 

% \TODO{[explain more in supp? ]} 
For edges, component entanglement often happens, when the model fails to determine whether the change is caused by reflectance or shading. Edges can result from the change of reflectance only, shading only or both. To evaluate the predictions, we need to have better understanding on what happens around these edges. 
In addition to edges with abrupt local changes, we're also interested in how well the prediction fits the groundtruth around the relatively flat \emph{regions}, e.g. in a constant-reflectance region, reflectance value shouldn't change while shading values should fit tightly to the changes. 

With the analysis above, we need to first design a data annotation scheme to annotate edges and regions of interest (Sec.~\ref{sec:annotation}). 
On top of the annotation, we design an diagnostic edge-aware metric to maximally reflect how model behaves (Sec.~\ref{sec:metric_design}). 
% both edge-based metrics and region-based metrics. 
% To improve the interpretability of evaluation metrics, we design a data annotation scheme and an edge-aware metric. 
% The common artifacts mainly lie in two categories: component entanglement and global inconsistency. 
% Component entanglement often happens on edges, when the decomposition fails to determine whether the change is caused by reflectance or shading. 
% For distant pixels that should have similar reflectance or shading, global inconsistency often happens due to the scale ambiguity of intrinsic images. 

\begin{figure}[t]
\centering
\includegraphics[width=\linewidth]{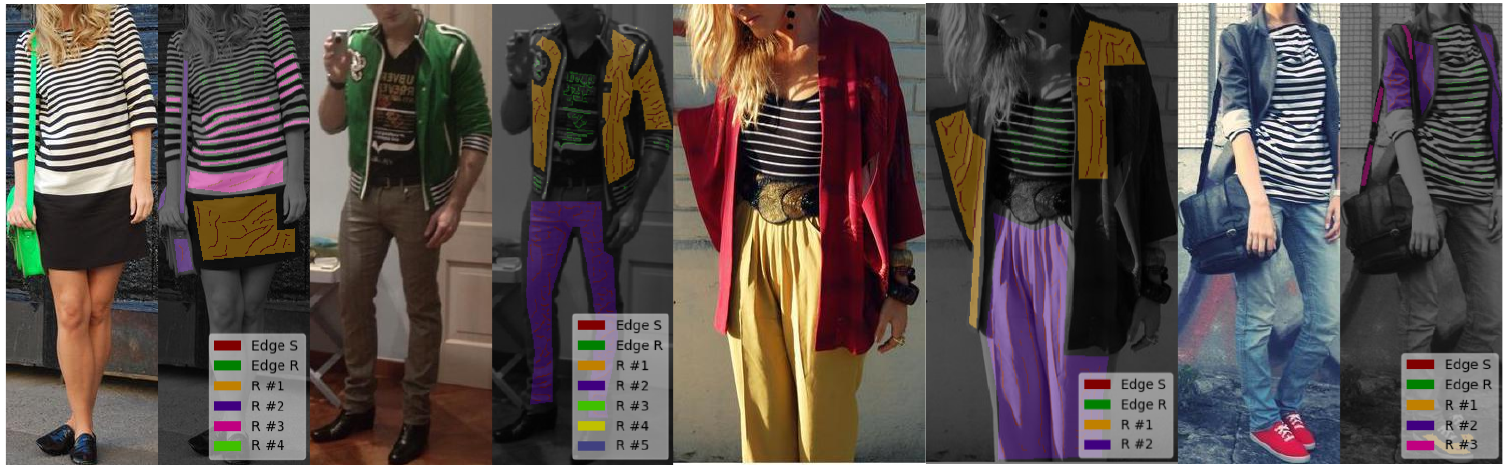} 
\caption{\footnotesize\textbf{Visualization of Annotation Result. } 
The left is RGB image and the right is the annotation results visualized on grayscale image. 
Red indicates edges caused only by shading ($\mathcal{E}_\text{S}$) and dark green shows edges caused only by reflectance ($\mathcal{E}_\text{R}$). Other colors indicate constant-reflectance regions $\{\mathcal{R}_c\}$. 
Best viewed zooming in. 
}
\label{fig:annotation}
\end{figure}

\subsection{Data Annotation Scheme}
\label{sec:annotation}
% As discussed above, to better evaluate the common artifacts of intrinsic models,
For CloIntrinsics testing set, we need data annotation on both edges $\mathcal{E}$ and regions $\mathcal{R}$ of interest. % $\{\mathcal{E}, \mathcal{R}\}$. 
Considering the difficulty and accuracy during annotation, we propose to focus on annotating constant-reflectance regions and edges caused by reflectance only or shading only. 

Specifically, region annotation $\mathcal{R} = \{\mathcal{R}_c | c=1,2,\cdots, C\}$ contains several regions $\mathcal{R}_c$ and elements in $\mathcal{R}_c$ have the same reflectance $c$. The annotators are asked to find pixels with intrinsically the same ``color'', i.e. the same reflectance, and annotate them with points or polygons. Instead of including comparison with inequal reflectance values, annotators can easily annotate more precise results. 

Edge annotation $\mathcal{E} = \{\mathcal{E}_\text{R}, \mathcal{E}_\text{S} \}$ is a subset of $\mathcal{E}_\text{Canny}$, which is edges detected by Canny algorithm \cite{canny1986computational}. 
$\mathcal{E}_\text{R}, \mathcal{E}_\text{S}$ denotes the edges caused by reflectance only and shading only, respectively. 
As we can use $\mathcal{E}_\text{S} = \mathcal{E}_\text{Canny} \cap \mathcal{R}$ to get the edges caused by shading, where $\mathcal{R}$ is the constant-reflectance regions we already annotated, 
we can simplify the annotation process and only need to annotate the edges caused by reflectance $\mathcal{E}_\text{R}$. 
To annotate $\mathcal{E}_\text{R}$, we design a user interface which shows Canny edges, and ask the annotators to add scribbles to the edges caused only by reflectance. 

% To make the edge annotation more precise, we implement a optimization-based algorithm based on the framework of \cite{zhao2012closed}. It uses Retinex constraint \cite{land1971lightness1} and enforce shading image to be smooth around annotated $\mathcal{E}_\text{R}$. 
% The decomposition results are displayed to the annotators, giving them a chance to make sure the annotation can lead to locally satisfying results. 
% Some annotation examples are shown in Fig.~\ref{fig:annotation}. 
The annotation collection pipeline includes data annotation and data verification, conducted by three annotators with background in intrinsic images. 
For data annotation, we ask two of the annotators to annotate the images twice. They are asked to only annotate the confident parts. 
The annotation will then be combined, and another inspector will verify the annotation. 
We noticed that it's hard for the inspector to find mistakes directly by watching the edge annotation, but it's easier for the inspector to judge whether reflectance and shading are well behaved. 
Based on this observation, we incorporate an interactive method \cite{bousseau2009user} for annotation inspection. 
Specifically, We use our edge annotation as the prior to solve for intrinsic images, and the results will be displayed to the annotator. 
Only when our annotation leads to plausible decomposition results can we assume the annotation is reliable. 
With the above professional annotation procedure, we obtained high quality annotation on real-world images.

\subsection{Metric Design}
\label{sec:metric_design}
With region annotation $\{\mathcal{R}_c | c=1,2,\cdots, C\}$ and edge annotation $\{\mathcal{E}_\text{R}, \mathcal{E}_\text{S} \}$, we define metric for region performance and edge performance separately. 

\noindent\textbf{Region reflectance metric. }
Pixels with the same reflectance should have similar values in the decomposition results. 
Given the predicted reflectance $\hat{R}$, we calculate the variance within each annotated region as a metric to evaluate the performance of predicted reflectance. 
Specifically, for each region $\mathcal{R}_c$, we first normalize the mean to $1$ considering the ambiguity in scale, 
\begin{equation}
	\tilde{R}_{i, j} = \frac{|\mathcal{R}_c|}{\sum_{(m, n) \in \mathcal{R}_c} \hat{R}_{m, n}} \ \hat{R}_{i, j}. 
	\label{eq:norm}
\end{equation}
Then we compute the variance as the error measure of reflectance
% \begin{equation}
% 	Var(\mathcal{R}_c) = \frac{1}{|\mathcal{R}_c|} \sum_{(i, j) \in \mathcal{R}_c} (\tilde{R}_{i, j} - 1)^2,
% \end{equation}
\begin{equation}
	\text{RegionError}_\text{R} = \frac 1C \sum_{c=1}^{C} \frac{1}{|\mathcal{R}_c|} \sum_{(i, j) \in \mathcal{R}_c} (\tilde{R}_{i, j} - 1)^2. 
\end{equation}

\noindent\textbf{Region shading metric. }
Within each constant reflectance region $\mathcal{R}_c$, we can assume the reflectance is constant and compute the estimated groundtruth shading $S_c$, which is RGB image divided by reflectance. 
% $\hat{S}_{i, j}\frac{I_{i, j}}{\hat{R}_{i, j}}, (i, j) \in \mathcal{R}_c$
Given the predicted shading $\hat{S}$, we compute scale-invariant mean square error (si-MSE) as a metric to evaluate global shading performance, 
\begin{equation}
	\text{si-MSE}_{\mathcal{R}}(\hat{X}, X) = \frac{1}{|\mathcal{R}|} \sum_{(i, j) \in \mathcal{R}} ||\alpha\hat{X}_{i,j} - X_{i,j}||^2,
\end{equation}
where $\alpha = \arg\min \sum_{(i, j) \in \mathcal{R}} ||\alpha\hat{X}_{i,j} - X_{i,j}||^2$, and
\begin{equation}
	\text{RegionError}_\text{S} = \frac{\sum_{c=1}^{C} \text{si-MSE}_{\mathcal{R}_c}(\hat{S}, S_c)}{\sum_{c=1}^{C} \text{si-MSE}_{\mathcal{R}_c}(\mathbf{0}, S_c) }
\end{equation}
where we use a relative error similar to \cite{grosse2009ground}. 
Considering that the estimated shading $S_c$ is not exactly the ground truth, we don't accumulate the part of error when the prediction error is within the range of $\pm 5\%$. 

\noindent\textbf{Edge metric. }
The most common artifact for edges is component entanglement. 
With edge annotation $\{\mathcal{E}_\text{R}, \mathcal{E}_\text{S}\}$, we're able to evaluate model performance in an more interpretable way. 
Ideally, the predicted reflectance $\hat{R}$ should change around $\mathcal{E}_\text{R}$ but not $\mathcal{E}_\text{S}$. 
Similarly, the predicted shading $\hat{S}$ should change around $\mathcal{E}_\text{S}$ but not $\mathcal{E}_\text{R}$. 
We use local gradient magnitude to indicate the local changes, and assume it changes when local gradient magnitude is larger than a threshold $\tau$. 
We first normalize in the same way as in Eq.~\ref{eq:norm} and get $\tilde{R}$ and $\tilde{S}$. 
Then, to indicate how well the predicted results agrees with the annotation, we define the accuracy of reflectance at $\mathcal{E}_\text{S}$ and the accuracy of shading at $\mathcal{E}_\text{R}$ as follows
\begin{align}
	\text{Acc}_\text{R}^{(\mathcal{E}_\text{S})} &= 
	\frac{\sum_{(i, j) \in \mathcal{E}_\text{S}} \mathbbm{1}\big( \|\nabla \tilde{R}_{i, j}\| < \tau \big) }{|\mathcal{E}_\text{S}|}, \\
	\text{Acc}_\text{S}^{(\mathcal{E}_\text{R})} &=
	\frac{\sum_{(i, j) \in \mathcal{E}_\text{R}} \mathbbm{1}\big( \|\nabla \tilde{S}_{i, j}\| < \tau \big) }{|\mathcal{E}_\text{R}|},
\end{align}
where $\mathbbm{1}(\cdot)$ is the indicator function. 
The accuracies defined above directly indicate how reflectance and shading entangle with each other, making it interpretable and diagnostic. 
Similarly, we can also define the accuracy of reflectance at $\mathcal{E}_\text{R}$ as $\text{Acc}_\text{R}^{(\mathcal{E}_\text{R})}$, and the accuracy of shading at $\mathcal{E}_\text{S}$ as $\text{Acc}_\text{S}^{(\mathcal{E}_\text{S})}$, 
\begin{align}
	\text{Acc}_\text{R}^{(\mathcal{E}_\text{R})} &= 
	\frac{\sum_{(i, j) \in \mathcal{E}_\text{R}} \mathbbm{1}\big( ||\nabla \tilde{R}_{i, j}|| > \tau \big) }{|\mathcal{E}_\text{R}|}, \\ 
	\text{Acc}_\text{S}^{(\mathcal{E}_\text{S})} &= 
	\frac{\sum_{(i, j) \in \mathcal{E}_\text{S}} \mathbbm{1}\big( ||\nabla \tilde{S}_{i, j}|| > \tau \big) }{|\mathcal{E}_\text{S}|}.
\end{align}
We finally use weighted harmonic mean as the metric for edge performance, 
\begin{align}
	F_\text{R} &= \frac{w_1 + w_2}{ \frac{w_1}{\text{Acc}_\text{R}^{(\mathcal{E}_\text{S})}} + \frac{w_2}{\text{Acc}_\text{R}^{(\mathcal{E}_\text{R})}}}, \\ 
	F_\text{S} &= \frac{w_1 + w_2}{ \frac{w_1}{\text{Acc}_\text{S}^{(\mathcal{E}_\text{R})}} + \frac{w_2}{\text{Acc}_\text{S}^{(\mathcal{E}_\text{S})}}}, 
	\label{eq:edge_metric}
\end{align}
% where $F_\text{R}$ and $F_\text{S}$ 
indicating edge performance of reflectance and shading respectively.

In summary, for the real-world testing set, we use $\text{RegionError}_\text{R}$, $\text{RegionError}_\text{S}$ to evaluate generally how well the predicted intrinsic images fit to the groundtruth. 
For common component entanglement artifacts, we use $\text{F}_\text{R}$ and $\text{F}_\text{S}$ to evaluate the edge performance of intrinsic models. By looking at $\text{Acc}_\text{R}^{(\mathcal{E}_\text{R})}, \text{Acc}_\text{R}^{(\mathcal{E}_\text{S})}, \text{Acc}_\text{S}^{(\mathcal{E}_\text{R})}, \text{Acc}_\text{S}^{(\mathcal{E}_\text{S})}$, we can learn diagnostic information on how model behaves around edges. 
% \TODO{more intuition here!!}
%!TEX root=main.tex

\begin{figure*}[t!]
\centering
\begin{minipage}[t]{0.45\textwidth}
\centering
\includegraphics[width=\textwidth]{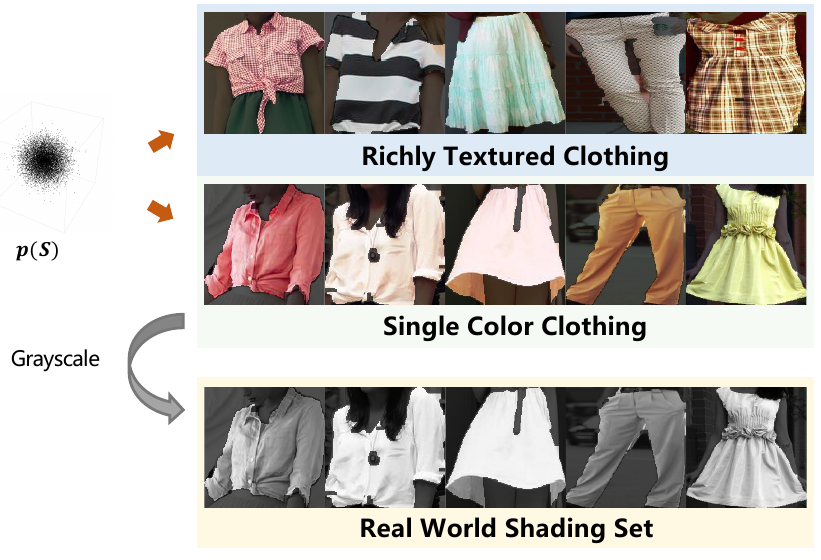}
\caption{\small \textbf{Visualization of Real World Shading Set.} 
Sharing similar clothing types and deformations, richly-textured and single-color clothing have similar shading distribution. 
We collect the shading of single-color clothing to form a real world shading set. 
}
\label{fig:shadingset}
\end{minipage}
\hspace{2mm}
\begin{minipage}[t]{0.50\textwidth}
\centering
\includegraphics[width=\textwidth]{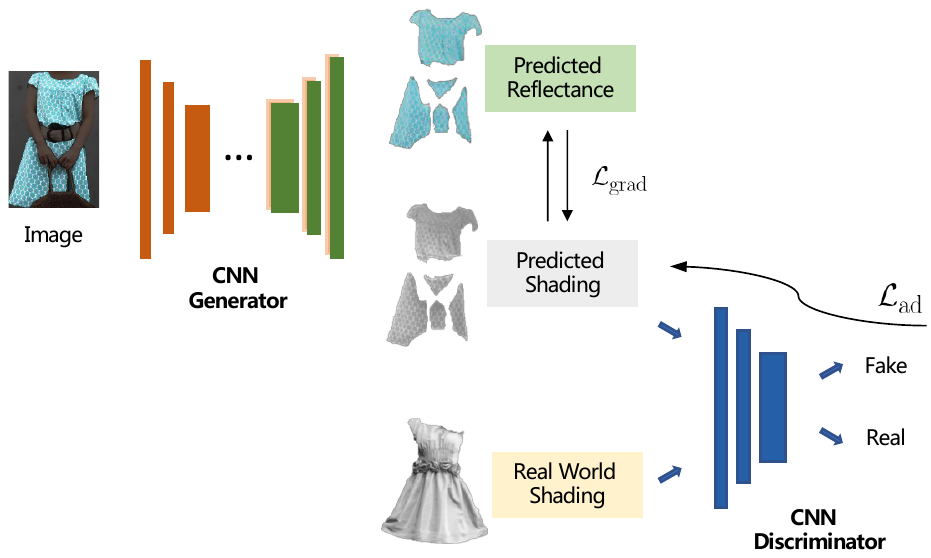}
\vspace{-3mm}
\caption{\small \textbf{Architecture of ClothInNet.} 
The CNN generator predicts reflectance and shading, and learn the real world shading distribution in an adversarial manner. 
Gradient constraint loss $\mathcal{L}_\text{grad}$ restricts simultaneous shading and reflectance changes.  
}
\label{fig:clothinnet}
\end{minipage}
\vspace{-4mm}
\end{figure*}

\vspace{-3mm}
\section{ClothInNet model} 
\label{sec:method}
\vspace{-1mm}

The bottleneck for learning-based approaches with CNNs is the lack of \emph{effective} data. 
Synthetic images lack the complexity of images in the wild and there's natually a domain gap, while annotations on real world images are so sparse that it's not adequate to learn a model in a purely data-driven manner. 
Existing datasets actually fail to provide the real world distribution of intrinsic images. 

To address this problem, we propose Clothing Intrinsics Network (ClothInNet) model. 
As we can note, the domain gap of shading images between synthetic and real world is much larger than reflectance images, due to the complex shape and illumination conditions. 
% The general idea of our method is
We propose to acquire real world shading from single-color clothing images with minor additional annotation. 
With these data, we can jointly train a CNN with an adversarial learning module. By discriminating the output shading from real world shading, the model learns to predict better shading results. 
Additionally, we introduce the gradient constraint loss as a prior for predicted reflectance and shading, to further improve the predictions. 

\vspace{-3mm}
\subsection{Real World Shading from Single-color Clothing}
\label{sec:shading_acquisition}
\vspace{-1mm}
As discussed above, real world shading data is vital but hard to densely annotate in pair. 
However, we can still make use of unpaired data. %with only minor additional annotation. 

As we can observe in fashion datasets like HumanParsing \cite{ATR,CO-CNN}, clothing can be roughly divided into two categories based on the differences in reflectance: \emph{single-color} clothing and \emph{richly-textured} clothing. 
As shown in Fig.~\ref{fig:shadingset}, the clothing types, shape deformations and lighting conditions should be similar for the two categories. 
And from the physical meaning of reflectance and shading, we can assume the shading of single-color clothing and richly-textured clothing should have similar distribution.
Intrinsic image decomposition for richly-textured clothing is quite hard, but it's pretty easy to get the approximate shading of single-color clothing. This is because, single-color clothing has nearly constant reflectance and simply the grayscale image is a good approximation of shading. 
Single-color clothing takes a considerable part (around $20\%$) in fashion datasets, and there're segmentation results of clothing items. We only need to annotate whether a clothing item is single-color, and we can collect a set of real world shading $\mathcal{S}_\text{real}$. 

Specifically, we annotated HumanParsing \cite{ATR,CO-CNN} dataset. 
We load the masked clothing items in the dataset and annotate whether it is single-color. 
This is only a single label so it's pretty easy to annotate. And because we care more about the precision than recall of the annotation, we trained a simple CNN classifier to largely facilitate the annotation process. 
We got 20008 clothing items after area thresholding and filtering noisy items, and finally picked out 2602 single-color clothing items to form the shading set $\mathcal{S}_\text{real}$. 

% To collect the real world shading set $\mathcal{S}_\text{real}$ described in Sec.~\ref{sec:shading_acquisition}, we need to annotate whether a clothing image belongs to single-color clothing. 
% We annotated HumanParsing\cite{ATR} dataset. We got 20008 clothing items after area thresholding to filter noisy items, and finally picked out 2602 single-color clothing items to form the shading set $\mathcal{S}_\text{real}$. 

\vspace{-3mm}
\subsection{Model Architecture}
\vspace{-1mm}
% \INFO{1. network components; 2. data and loss}
ClothInNet contains a CNN generator and a CNN discriminator, as shown in Fig.~\ref{fig:clothinnet}. 
We train the model with CloIntrinsics $\{\mathcal{I}_\text{synthetic}, \mathcal{R}_\text{synthetic}, \mathcal{S}_\text{synthetic}\}$, a real world clothing image set $\{\mathcal{I}_\text{real}\}$, and the real world shading set $\{\mathcal{S}_\text{real}\}$ acquired in Sec.~\ref{sec:shading_acquisition}. 
% Please note that the synthetic dataset doesn't necessarily need to be in clothing domain. 

We use the synthetic dataset to produce dense supervision to the model, which helps it learn the basic principles of intrinsic images. 
The real world clothing image set $\{\mathcal{I}_\text{real}\}$ is to transfer the model to real world domain. 
The real world shading set $\{\mathcal{S}_\text{real}\}$ is used as positive sample to train an adversarial discriminator, and the CNN generator learns to fit to the real world distribution through adversarial training. 
Finally, the gradient constraint loss prevent reflectance and shading from changing simultaneously, which further constrain the texture-copying artifacts. 
We separately describe the supervision signals in the following part of this section. 

\vspace{1mm}
\noindent\textbf{Direct Supervision. }
Similar to previous models, the CNN generator consumes RGB images as input and predicts corresponding reflectance and shading images as outputs. 
When training with the synthetic dataset $\{\mathcal{I}_\text{synthetic}, \mathcal{R}_\text{synthetic}, \mathcal{S}_\text{synthetic}\}$, the model learns the general knowledge on intrinsic images from dense supervision. 
% which is
% \begin{equation}
% 	\text{si-MSE}(\hat{X}, X) = \frac{1}{N} \sum_{i=1}^N ||\alpha\hat{X}_i - X_i||^2,
% \end{equation}
% where $\alpha = \arg\min\sum_{i=1}^N ||\alpha\hat{X}_i - X_i||^2.$
We use scale-invariant mean square error loss (si-MSE) as supervision signal, and enforce the predicted intrinsic images and its first order gradient to match with ground truth. 
The loss function can be written as follows
\begin{equation}
	\mathcal{L}_\text{R} = \text{si-MSE}(\hat{R}, R) + \text{si-MSE}(\nabla \hat{R}, \nabla R), 
	\label{eq:regress_R}
\end{equation}
\begin{equation}
	\mathcal{L}_\text{S} = \text{si-MSE}(\hat{S}, S) + \text{si-MSE}(\nabla \hat{S}, \nabla S). 
	\label{eq:regress_S}
\end{equation}
% where $\lambda_{\text{grad}}$ is the ratio of loss terms. 
Additionally, we use reconstruction loss 
\begin{equation}
	\mathcal{L}_\text{reconstruct} = \frac{1}{N} \sum_{i=1}^N (I_i - R_i \cdot S_i)^2 
	\label{eq:reconstruct}
\end{equation}
to encourage Hadamard product relationship $I = R \odot S$. 

The direct supervision through synthetic dataset can form a multi-task loss
\begin{equation}
	\mathcal{L}_\text{direct} = \lambda_\text{R} \mathcal{L}_\text{R} + \lambda_\text{S} \mathcal{L}_\text{S} + \mathcal{L}_\text{reconstruct} 
\end{equation}
where $\lambda_\text{R}, \lambda_\text{S}$ weigh different terms.

\vspace{1mm}
\noindent\textbf{Adversarial Supervision. }
When training with real world images $\mathcal{I}_\text{real}$, there's no paired groundtruth. But we can train a discriminator and learn from $\mathcal{S}_\text{real}$ in an adversarial manner. 

The discriminator $D$ takes shading images $S$ as input and predicts a binary label indicating whether it's a real shading image. 
The positive samples come from the real world shading set $\{\mathcal{S}_\text{real}\}$ acquired in Sec.~\ref{sec:shading_acquisition}. 
The negative samples includes the shading images predicted by the CNN generator and the grayscale image of richly-textured clothing. 
The loss function to train the discriminator is
\begin{equation}
\mathcal{L}_{\text{D}} = \text{BCE}(D(S), \mathbbm{1}(S \in \mathcal{S}_\text{real}))
\label{eq:loss_D}
\end{equation}
where $\text{BCE}$ is binary cross entropy loss
\begin{equation}
	\text{BCE}(\hat{y}, y) = y\log\hat{y} + (1-y)\log(1-\hat{y}),
\end{equation}
and $\mathbbm{1}(\cdot)$ is the indicator function. 

To train the CNN generator to predict realistic shading that fools the discriminator, for the shading prediction $\hat{S}$, we define the adversarial loss
\begin{equation}
\mathcal{L}_{\text{ad}} = \text{BCE}(D(\hat{S}), 1)
\end{equation}
as additional supervision for CNN generator. 
From a probabilistic view, we know $p(S|I) \propto p(I|S)p(S).$
The shading distribution $p(S)$ provided by $\mathcal{S}_\text{real}$ is helpful for training models when there's no available pairwise dense annotation.

\vspace{1mm}
\noindent\textbf{Exclusive Gradient Constraint. } 
Large discontinuities seldom occur at the same time in reflectance and shading. And when this happens, there're usually texture-copying artifacts. 
As the adversarial supervision already improves the shading performance, we introduce a novel loss term
\begin{equation}
	\mathcal{L}_\text{grad} = \frac{1}{N} \sum_{i=1}^N ||\nabla \hat{R}_i \cdot \nabla \hat{S}_i||^2
\end{equation}
to further constrain the edge behaviour and improve the performance of reflectance. 
% This loss will only backpropagate to reflectance image. 
When shading is changing, the loss $\mathcal{L}_\text{grad}$ penalize the local changes of reflectance or at least in the different direction from $\nabla \hat{S}$. This simple loss term enforce explicit reasoning around edges, and brings amazing improvements on reflectance. 

% \vspace{1mm}
To sum up, we train our ClothInNet model with $\mathcal{L}_\text{D}$ in Eq.~\ref{eq:loss_D} for CNN discriminator and
\begin{equation}
	\mathcal{L}_\text{G} = \mathcal{L}_\text{direct} + \lambda_\text{ad} \mathcal{L}_{\text{ad}} + \lambda_\text{grad} \mathcal{L}_\text{grad}
\end{equation}
for CNN generator, where $\lambda_\text{ad}, \lambda_\text{grad}$ weigh the loss terms. 

% \subsection{Implementation Details}
% To collect the real world shading set $\mathcal{S}_\text{real}$ described in Sec.~\ref{sec:shading_acquisition}, we need to annotate whether a clothing image belongs to single-color clothing. 
% This is a pretty easy-to-annotate label, and because we only care about the precision (not recall) of the annotation, we can train a simple classifier to largely facilitate the annotation process. 
% We annotated HumanParsing\cite{ATR} dataset. We got 20008 clothing items after area thresholding to filter noisy items, and finally picked out 2602 single-color clothing items to form the shading set $\mathcal{S}_\text{real}$. 

% The architecture of CNN generator is UNet\cite{ronneberger2015u} with residual links\cite{he2016deep}. 
% The CNN discriminator contains three intermediate convolutional layers, with batch normalization\cite{ioffe2015batch} and LeakyReLU\cite{glorot2011deep} in each layer. 

%!TEX root=main.tex

\begin{figure*}[t!]
\centering
\vspace{-3mm}
\begin{tabular}{@{}c@{}c@{}c@{}c@{}c@{}c@{}c@{}c@{}c@{}c@{}c@{}}
\scalebox{1.0}[0.8]{\includegraphics[width=0.09\textwidth]{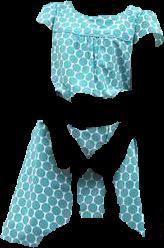}} & 
\scalebox{1.0}[0.8]{\includegraphics[width=0.09\textwidth]{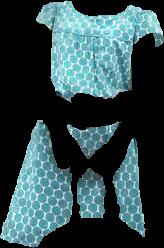}} & 
\scalebox{1.0}[0.8]{\includegraphics[width=0.09\textwidth]{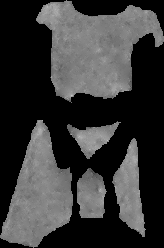}} & 
\scalebox{1.0}[0.8]{\includegraphics[width=0.09\textwidth]{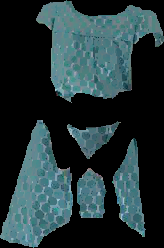}} & 
\scalebox{1.0}[0.8]{\includegraphics[width=0.09\textwidth]{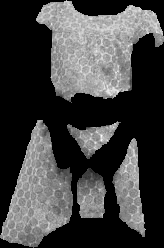}} & 
\scalebox{1.0}[0.8]{\includegraphics[width=0.09\textwidth]{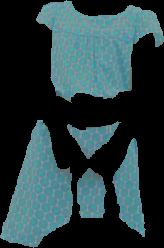}} & 
\scalebox{1.0}[0.8]{\includegraphics[width=0.09\textwidth]{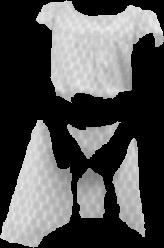}} & 
\scalebox{1.0}[0.8]{\includegraphics[width=0.09\textwidth]{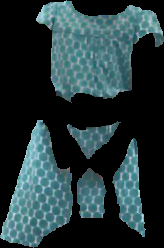}} & 
\scalebox{1.0}[0.8]{\includegraphics[width=0.09\textwidth]{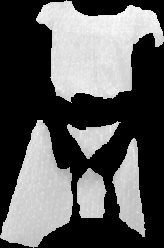}} & 
\scalebox{1.0}[0.8]{\includegraphics[width=0.09\textwidth]{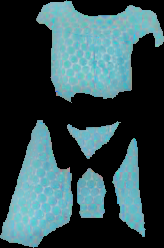}} & 
\scalebox{1.0}[0.8]{\includegraphics[width=0.09\textwidth]{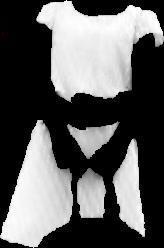}} \\
\scalebox{1.0}[0.8]{\includegraphics[width=0.09\textwidth]{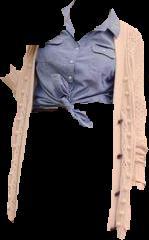}} & 
\scalebox{1.0}[0.8]{\includegraphics[width=0.09\textwidth]{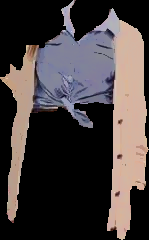}} & 
\scalebox{1.0}[0.8]{\includegraphics[width=0.09\textwidth]{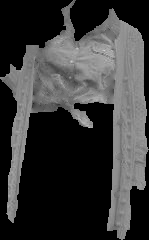}} & 
\scalebox{1.0}[0.8]{\includegraphics[width=0.09\textwidth]{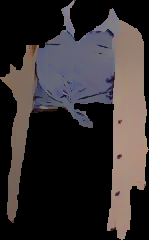}} & 
\scalebox{1.0}[0.8]{\includegraphics[width=0.09\textwidth]{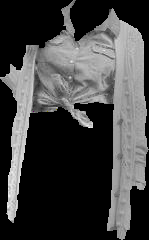}} & 
\scalebox{1.0}[0.8]{\includegraphics[width=0.09\textwidth]{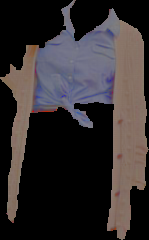}} & 
\scalebox{1.0}[0.8]{\includegraphics[width=0.09\textwidth]{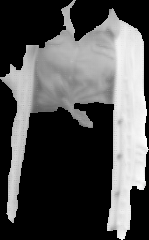}} & 
\scalebox{1.0}[0.8]{\includegraphics[width=0.09\textwidth]{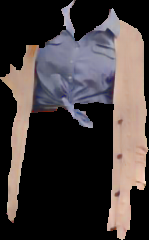}} & 
\scalebox{1.0}[0.8]{\includegraphics[width=0.09\textwidth]{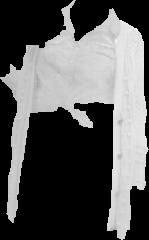}} & 
\scalebox{1.0}[0.8]{\includegraphics[width=0.09\textwidth]{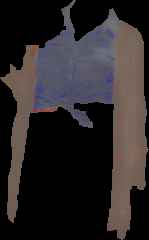}} & 
\scalebox{1.0}[0.8]{\includegraphics[width=0.09\textwidth]{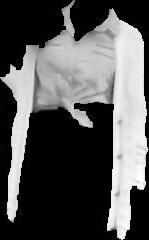}} \\
\scalebox{1.0}[0.6]{\includegraphics[width=0.09\textwidth]{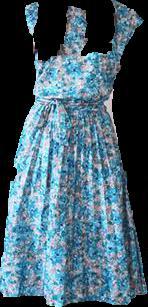}} & 
\scalebox{1.0}[0.6]{\includegraphics[width=0.09\textwidth]{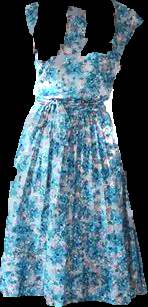}} & 
\scalebox{1.0}[0.6]{\includegraphics[width=0.09\textwidth]{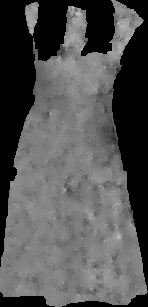}} & 
\scalebox{1.0}[0.6]{\includegraphics[width=0.09\textwidth]{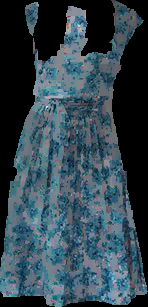}} & 
\scalebox{1.0}[0.6]{\includegraphics[width=0.09\textwidth]{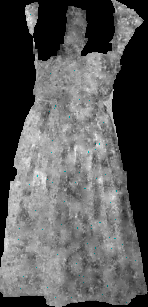}} & 
\scalebox{1.0}[0.6]{\includegraphics[width=0.09\textwidth]{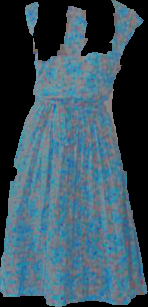}} & 
\scalebox{1.0}[0.6]{\includegraphics[width=0.09\textwidth]{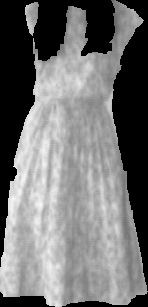}} & 
\scalebox{1.0}[0.6]{\includegraphics[width=0.09\textwidth]{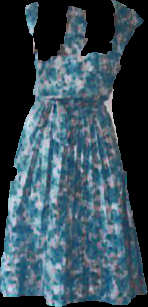}} & 
\scalebox{1.0}[0.6]{\includegraphics[width=0.09\textwidth]{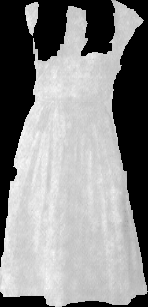}} & 
\scalebox{1.0}[0.6]{\includegraphics[width=0.09\textwidth]{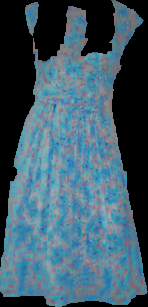}} & 
\scalebox{1.0}[0.6]{\includegraphics[width=0.09\textwidth]{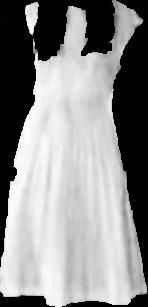}} \\
\scalebox{1.0}[0.8]{\includegraphics[width=0.09\textwidth]{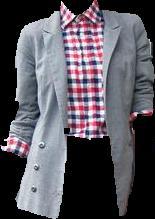}} & 
\scalebox{1.0}[0.8]{\includegraphics[width=0.09\textwidth]{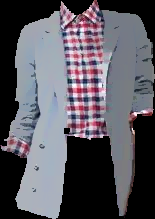}} & 
\scalebox{1.0}[0.8]{\includegraphics[width=0.09\textwidth]{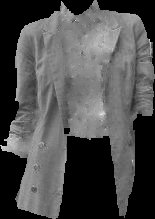}} & 
\scalebox{1.0}[0.8]{\includegraphics[width=0.09\textwidth]{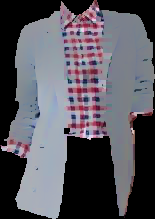}} & 
\scalebox{1.0}[0.8]{\includegraphics[width=0.09\textwidth]{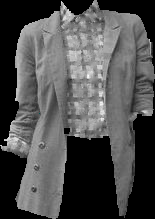}} & 
\scalebox{1.0}[0.8]{\includegraphics[width=0.09\textwidth]{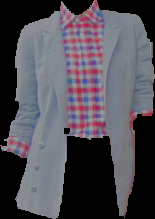}} & 
\scalebox{1.0}[0.8]{\includegraphics[width=0.09\textwidth]{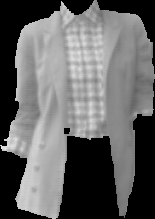}} & 
\scalebox{1.0}[0.8]{\includegraphics[width=0.09\textwidth]{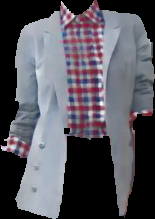}} & 
\scalebox{1.0}[0.8]{\includegraphics[width=0.09\textwidth]{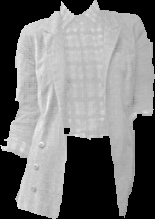}} & 
\scalebox{1.0}[0.8]{\includegraphics[width=0.09\textwidth]{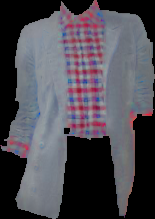}} & 
\scalebox{1.0}[0.8]{\includegraphics[width=0.09\textwidth]{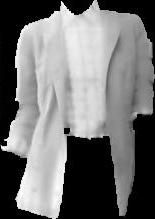}} \\
\scalebox{1.0}[0.8]{\includegraphics[width=0.09\textwidth]{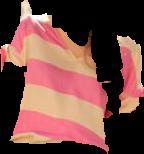}} & 
\scalebox{1.0}[0.8]{\includegraphics[width=0.09\textwidth]{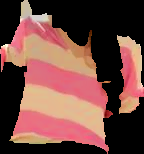}} & 
\scalebox{1.0}[0.8]{\includegraphics[width=0.09\textwidth]{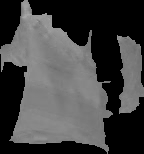}} & 
\scalebox{1.0}[0.8]{\includegraphics[width=0.09\textwidth]{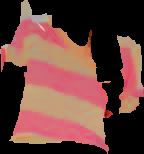}} & 
\scalebox{1.0}[0.8]{\includegraphics[width=0.09\textwidth]{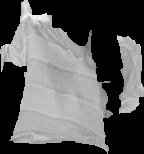}} & 
\scalebox{1.0}[0.8]{\includegraphics[width=0.09\textwidth]{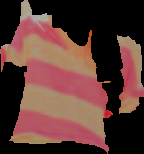}} & 
\scalebox{1.0}[0.8]{\includegraphics[width=0.09\textwidth]{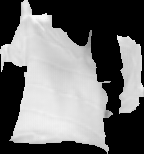}} & 
\scalebox{1.0}[0.8]{\includegraphics[width=0.09\textwidth]{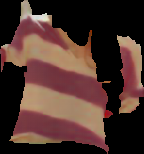}} & 
\scalebox{1.0}[0.8]{\includegraphics[width=0.09\textwidth]{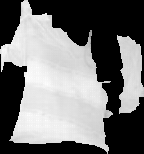}} & 
\scalebox{1.0}[0.8]{\includegraphics[width=0.09\textwidth]{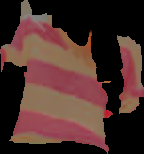}} & 
\scalebox{1.0}[0.8]{\includegraphics[width=0.09\textwidth]{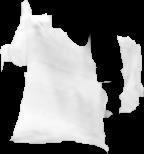}} \\
\scalebox{1.0}[0.8]{\includegraphics[width=0.09\textwidth]{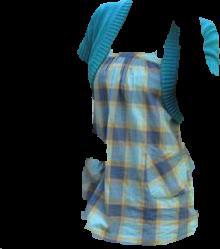}} & 
\scalebox{1.0}[0.8]{\includegraphics[width=0.09\textwidth]{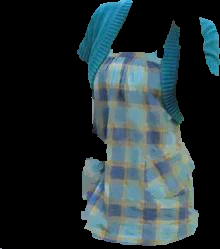}} & 
\scalebox{1.0}[0.8]{\includegraphics[width=0.09\textwidth]{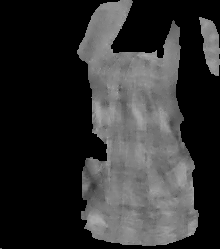}} & 
\scalebox{1.0}[0.8]{\includegraphics[width=0.09\textwidth]{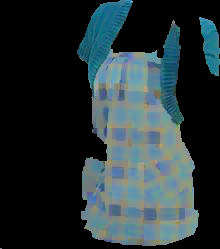}} & 
\scalebox{1.0}[0.8]{\includegraphics[width=0.09\textwidth]{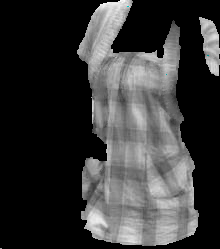}} & 
\scalebox{1.0}[0.8]{\includegraphics[width=0.09\textwidth]{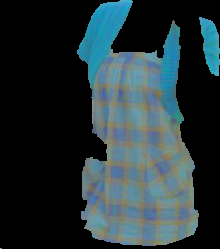}} & 
\scalebox{1.0}[0.8]{\includegraphics[width=0.09\textwidth]{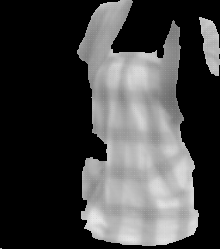}} & 
\scalebox{1.0}[0.8]{\includegraphics[width=0.09\textwidth]{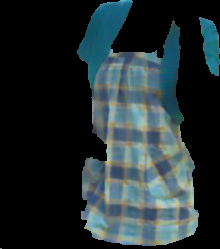}} & 
\scalebox{1.0}[0.8]{\includegraphics[width=0.09\textwidth]{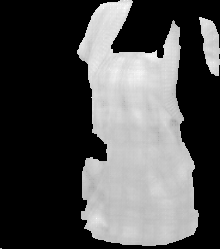}} & 
\scalebox{1.0}[0.8]{\includegraphics[width=0.09\textwidth]{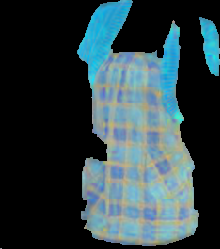}} & 
\scalebox{1.0}[0.8]{\includegraphics[width=0.09\textwidth]{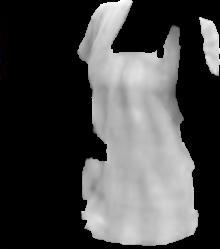}} \\
\scalebox{1.0}[0.8]{\includegraphics[width=0.09\textwidth]{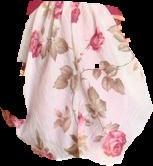}} & 
\scalebox{1.0}[0.8]{\includegraphics[width=0.09\textwidth]{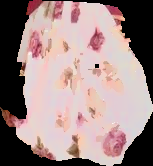}} & 
\scalebox{1.0}[0.8]{\includegraphics[width=0.09\textwidth]{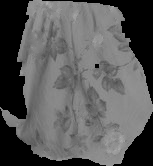}} & 
\scalebox{1.0}[0.8]{\includegraphics[width=0.09\textwidth]{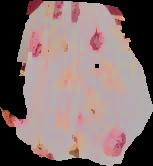}} & 
\scalebox{1.0}[0.8]{\includegraphics[width=0.09\textwidth]{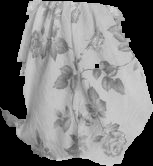}} & 
\scalebox{1.0}[0.8]{\includegraphics[width=0.09\textwidth]{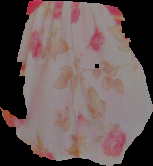}} & 
\scalebox{1.0}[0.8]{\includegraphics[width=0.09\textwidth]{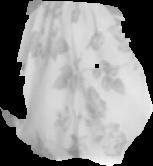}} & 
\scalebox{1.0}[0.8]{\includegraphics[width=0.09\textwidth]{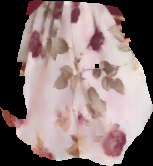}} & 
\scalebox{1.0}[0.8]{\includegraphics[width=0.09\textwidth]{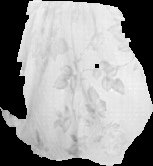}} & 
\scalebox{1.0}[0.8]{\includegraphics[width=0.09\textwidth]{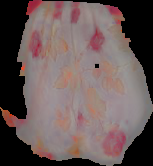}} & 
\scalebox{1.0}[0.8]{\includegraphics[width=0.09\textwidth]{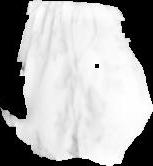}} \\

{\scriptsize Image } & 
\multicolumn{2}{c}{\scriptsize Retinex \cite{land1971lightness1}} & 
\multicolumn{2}{c}{\scriptsize Zhao et al. \cite{zhao2012closed}} & 
\multicolumn{2}{c}{\scriptsize BigTime \cite{li2018learning}} & 
\multicolumn{2}{c}{\scriptsize CGIntrinsics \cite{li2018cgintrinsics}} & 
\multicolumn{2}{c}{\scriptsize ClothInNet} \\
\end{tabular}
\vspace{-3mm}
\caption{\footnotesize \textbf{Qualitative comparisons on real world clothing images. } 
For each method, the left result is reflectance and the right is shading. 
Our model significantly removes texture-copying artifacts while retaining tiny details. 
% * indicates the model is trained on CGI, IIW and SAW. 
Best viewed zooming in.
}
\label{fig:qualitative} 
\vspace{-3mm}
\end{figure*}

\vspace{-4mm}
\section{Experiments}
\label{sec:results}
\vspace{-2mm}

In this section, we perform extensive evaluation on our methods, including qualitative results on real world clothing images and quantitative comparison with state-of-the-art methods. 
We quantitatively evaluate the methods on CloIntrinsics testing set and provide additional cross-domain evaluation on MIT Intrinsic Images Dataset \cite{grosse2009ground}. 
We first introduce detail settings in Sec.~\ref{sec:settings}. Then we compare qualitatively and quantitatively in Sec.~\ref{sec:qualitative} and Sec.~\ref{sec:quantitative}. We further show some graphics applications in Sec.~\ref{sec:app}. 

\vspace{-3mm}
\subsection{Experiment Settings}
\vspace{-1mm}
\label{sec:settings}
\noindent\textbf{Implementation details.}
% The CNN generator is UNet\cite{ronneberger2015u} with residual connections\cite{he2016deep}. 
The architecture of CNN generator is UNet \cite{ronneberger2015u} with residual connections \cite{he2016deep}. 
The CNN discriminator contains three intermediate convolutional layers, with batch normalization \cite{ioffe2015batch} and LeakyReLU \cite{glorot2011deep} in each layer. 
The real world shading image set $\{\mathcal{S}_\text{real}\}$ used during training contains 2602 single-color clothing images from HumanParsing dataset \cite{ATR,CO-CNN}, and is annotated as described in Sec.~\ref{sec:shading_acquisition}. 
The real world clothing image set $\{\mathcal{I}_\text{real}\}$ is a subset of HumanParsing dataset. We cropped out the clothing images with label \emph{upper-clothes, skirt, pants, dress, hat, bag, scarf} and used area thresholding to filter noisy images. 
% The synthetic dataset $\{\mathcal{I}_\text{synthetic}, \mathcal{R}_\text{synthetic}, \mathcal{S}_\text{synthetic}\}$ to train ClothInNet is either CGIntrinsics dataset \cite{li2018cgintrinsics} or CloIntrinsics training set. 
The loss ratios are $\lambda_\text{R} = \lambda_\text{S} = 1$ and $ \lambda_\text{ad} = \lambda_\text{grad} = 0.1$. 
The input resolution is 256x256 and the batch size is 8. The model is trained for 80 epochs with learning rate as 3e-4. 

For edge-aware metric of CloIntrinsics testing set, we set the weights $w_1=3, w_2=1$ in Eq.~\ref{eq:edge_metric}. 
This allows $F_\text{R}$ and $F_\text{S}$ to focus more on the entangled edges. 
This won't reduce the strictness of evaluation, as the less-weighted $\text{Acc}_\text{R}^{(\mathcal{E}_\text{R})}$ and $\text{Acc}_\text{S}^{(\mathcal{E}_\text{S})}$ will also reflect in region metrics. 

\vspace{1mm}
\noindent\textbf{Models and baselines.} 
We compare our ClothInNet model with two representative optimization-based methods, including Retinex algorithm \cite{land1971lightness1} and Zhao et al. \cite{zhao2012closed}, and two state-of-the-art deep learning methods BigTime \cite{li2018learning} and CGIntrinsics \cite{li2018cgintrinsics}. 
% BigTime \cite{li2018learning} trains with image sequences of same scenes under different lighting conditions. As the dataset contains both indoor and outdoor scenes, BigTime generalize well to challenging cases. 
% CGIntrinsics \cite{li2018cgintrinsics} proposed a largescale synthetic scene dataset and trained a UNet \cite{ronneberger2015u} with these data. 
% BigTime and CGIntrinsics are among current state-of-the-art methods and will be 
% We compare with the two methods as they represent current state-of-the-art. 
In the ablation study, the baseline method is to train a UNet with only regression loss and reconstruction loss (Eq.~\ref{eq:regress_R}, ~\ref{eq:regress_S}, ~\ref{eq:reconstruct}) on the CloIntrinsics training set. We will refer to this baseline as UNet-CLO. 

% To validate the effectiveness of both our ClothInNet model and CloIntrinsics data, we separately train ClothInNet with CGIntrinsics and CloIntrinsics and will be referred to as \emph{ClothInNet-CGI} and \emph{ClothInNet-CLO}. 

% Our proposed method with adversarial module on top of UNet-baseline and UNet-CLO will be refered to as \emph{Ours-CGI} and \emph{Ours-CLO}. 
% To fairly compare with this model in clothing domain, inspired by \cite{janner2017self}, a stronger baseline is to simultaneously use reconstruction loss on real world clothing images (HumanParsing\cite{ATR} dataset in experiments) in addition to training with CGIntrinsics data. 
% We will refer to this baseline method as \emph{UNet-CGI}. 
% To test our proposed method, we train our method with CGIntrinsics and HumanParsing\cite{ATR} dataset 

\begin{table*}[t!]
\vspace{-1mm}
\begin{center}
\caption{\small Quantitative evaluation on CloIntrinsics testing set. }
\label{tbl:clointrinsics_quantitative} 
\addtolength{\tabcolsep}{3pt}
\resizebox{\textwidth}{!}{
\begin{tabular}{|l|c|c|c|c|c|c|c|c|}
\hline
Model & $\text{Acc}_\text{R}^{(\mathcal{E}_\text{S})}$ & $\text{Acc}_\text{R}^{(\mathcal{E}_\text{R})}$ & $\text{Acc}_\text{S}^{(\mathcal{E}_\text{R})}$ & $\text{Acc}_\text{S}^{(\mathcal{E}_\text{S})}$ & $F_\text{R}$ & $F_\text{S}$ & $\text{RegionError}_\text{R}$ & $\text{RegionError}_\text{S}$ \\
\hline 
% UNet-CGI  			& 0.5938	& 0.9875	& 0.5351	& \textbf{0.9606}	& 0.6595	& 0.6017	& 0.0155	& 0.0241	\\
% + Ad Supervision  	& 0.5983	& \textbf{0.9879}	& 0.7303	& 0.7763	& 0.6637	& 0.7413	& 0.0144	& 0.0225	\\
% + Grad Constraint  	& \textbf{0.6474}	& 0.9736	& \textbf{0.7481}	& 0.7652	& \textbf{0.7066}	& \textbf{0.7523}	& \textbf{0.0131}	& \textbf{0.0214}	\\
% \hline 
% UNet-CLO  			& 0.6601	& 0.9656	& 0.5756	& \textbf{0.9233}	& 0.7168	& 0.6354	& 0.0136	& \textbf{0.0162}	\\
% + Ad Supervision  	& 0.6058	& \textbf{0.9903}	& 0.8714	& 0.6686	& 0.6709	& 0.8100	& 0.0133	& 0.0173	\\
% + Grad Constraint  	& \textbf{0.6737}	& 0.9756	& \textbf{0.8816}	& 0.6813	& \textbf{0.7303}	& \textbf{0.8212}	& \textbf{0.0109}	& 0.0167	\\
Retinex \cite{land1971lightness1} & 0.3890 & 0.9209 & 0.4488 & 0.9346 & 0.4546 & 0.5158 & 0.0383 & 0.0152 \\
Zhao et al. \cite{zhao2012closed} & 0.5373 & 0.8874 & 0.3366 & 0.9915 & 0.5961 & 0.4032 & 0.0434 & 0.0129 \\
BigTime \cite{li2018learning} & 0.5646 & 0.9952 & 0.5633 & \textbf{0.9968} & 0.6331 & 0.6320 & \textbf{0.0039} & 0.0115 \\
CGIntrinsics \cite{li2018cgintrinsics} & 0.4450 & 0.9945 & 0.7392 & 0.9317 & 0.5163 & 0.7794 & 0.0159 & 0.0193 \\
\hline 
% UNet-CGI & 0.4379 & \textbf{0.9985} & 0.5223 & 0.9456 & 0.5093 & 0.5881 & 0.0115 & 0.0200 \\
UNet-CLO & 0.5643 & 0.9846 & 0.4570 & 0.8705 & 0.6317 & 0.5185 & 0.0171 & 0.0084 \\
+ Ad Supervision & 0.6081 & \textbf{0.9962} & 0.7962 & 0.9373 & 0.6737 & 0.8273 & 0.0169 & 0.0089 \\
+ Grad Constraint & \textbf{0.7780} & 0.9943 & \textbf{0.8172} & 0.9447 & \textbf{0.8227} & \textbf{0.8457} & 0.0055 & \textbf{0.0076} \\
\hline 
\end{tabular}
}
\end{center}
\vspace{-3mm}
\end{table*}

\begin{table}[t]
\footnotesize
\begin{center}
\vspace{-6mm}
\addtolength{\tabcolsep}{3pt}
\caption{\small Quantitative cross-domain evaluation on MIT dataset. }
\label{tbl:mit_quantitative} 
\resizebox{0.45\textwidth}{!}{
\begin{tabular}{|l|c|c|}
\hline
Model & LMSE - R & LMSE - S \\
\hline 
Retinex \cite{land1971lightness1}	& 0.0366	& 0.0419	\\
Zhao et al. \cite{zhao2012closed}	& \textbf{0.0311}	& 0.0267	\\
BigTime \cite{li2018learning}	& 0.0341	& 0.0253	\\
CGIntrinsics \cite{li2018cgintrinsics}	& 0.0349	& 0.0259	\\
% ClothInNet-CGI  		& 0.0365	& 0.0266	\\
ClothInNet	  	& 0.0357	& \textbf{0.0229}	\\
\hline
\end{tabular}
}
\end{center}
\vspace{-9mm}
\end{table}

\vspace{-4mm}
\subsection{Qualitative Comparison}
\vspace{-1mm}
\label{sec:qualitative}
We show qualitative results on real world clothing images in Fig.~\ref{fig:qualitative}. 
Clothing images are challenging for intrinsic models as the intrinsic components have complex and entangled local changes. 
Traditional optimization based methods generally performs well around locally flat regions but completely fail when there're complex texture patterns. 
This is reasonable as traditional methods highly depend on hand-crafted priors, which can easily fail in wild images. 

Deep learning methods are generally more robust to noisy input images but still suffers from texture-copying artifacts. 
Take the first row as an example, the shading generated by BigTime \cite{li2018learning} produce severe noises around edges caused by reflectance changes. 
CGIntrinsics \cite{li2018cgintrinsics} relieves these noises but still cannot totally remove them. If we look into details of shading images predicted by CGIntrinsics, there're still noises around the edges. Also, CGIntrinsics sacrifice the details in shading images, losing the shading information in the original image. 
% Such intrinsic images can't be directly applied to real world applications. 
Our proposed ClothInNet model significantly improves the performance compared to these state-of-the-art methods and produces well-behaved intrinsic images. 
% As Fig.~\ref{fig:qualitative} shows, our proposed method works for both CGIntrinsics and CloIntrinsics dataset. 
In these challenging cases, ClothInNet significantly and stably reduces the texture-copying artifacts, and largely preserves the detailed information at the same time. For example, the shading of the richly-textured dress in the third row almost totally removes the surface texture, and it even preserves the tiny wrinkles in the middle. The predicted reflectance image also removes the edges caused by shading changes, significantly outperforming baseline methods. 

% Comparing ClothInNet-CGI with ClothInNet-CLO, we can see that model trained with CloIntrinsics generally produces better intrinsic images. 
% ClothInNet-CGI sometimes still fails to remove the surface texture while ClothInNet-CLO has a better performance. 
% Also, ClothInNet-CLO generally produces shading image with more details, making it fits better to the original image. 
% This validates the synthetic clothing data provides knowledge on diverse shapes and texture patterns in the clothing domain. 
% And thus the predicted intrinsic images, especially shading, tightly fit the groundtruth. 

% By comparing UNet-CGI, UNet-CLO with Ours-CGI and Ours-CLO, our proposed method effectively removes the texture-copying artifacts 
% For Ours-CGI, the generated shading tries hard to remove the reflectance component and the noises caused by $\mathcal{E}_\text{R}$ is smoothed compared to UNet-CGI. 
% With further help from CloIntrinsics data, Ours-CLO generates plausible shading image while preserving even tiny shape deformation in the original image, 

\vspace{-4mm}
\subsection{Quantitative Comparison}
\label{sec:quantitative}
\vspace{-2mm}

We conduct quantitative evaluation of baseline methods and ablation study for ClothInNet model on CloIntrinsics testing set. % including the adversarial module and gradient constraint. 
% We evaluate with CloIntrinsics testing set and our proposed edge-aware metric. 
Please note CloIntrinsics testing set contains real-world images and is independent from CloIntrinsics training set, so the comparison is fair. 
We use $F_\text{R}$, $F_\text{S}$ as the measure of edge performance and $\text{RegionError}_\text{R}$, $\text{RegionError}_\text{S}$ as metrics for general global performance. 
$\text{Acc}_\text{R}^{(\mathcal{E}_\text{S})}, \text{Acc}_\text{R}^{(\mathcal{E}_\text{R})}, \text{Acc}_\text{S}^{(\mathcal{E}_\text{R})}, \text{Acc}_\text{S}^{(\mathcal{E}_\text{S})}$ can give interpretable information on how model performs around the edges. 
The results are shown in Table \ref{tbl:clointrinsics_quantitative}. 

Our proposed metrics give novel insights on how model performs around edges and regions of interest. 
All baseline methods perform well on $\text{Acc}_\text{R}^{(\mathcal{E}_\text{R})}$ and $\text{Acc}_\text{S}^{(\mathcal{E}_\text{S})}$, but have rather low scores on $\text{Acc}_\text{R}^{(\mathcal{E}_\text{S})}$ and $\text{Acc}_\text{S}^{(\mathcal{E}_\text{R})}$. This indicates that the models learn to recall the local changes in the prediction but is careless in the precision. This leads to severe texture-copying artifacts. 
Our ClothInNet model significantly improves the quantitative performance and we show the ablation study in Table \ref{tbl:clointrinsics_quantitative}. 
We start with a UNet model trained on CloIntrinsics with direct supervision, which we refer to as UNet-CLO. 
Then we add our proposed adversarial module and gradient constraint loss on top. 
As we see, the adversarial module significantly improves the shading edge performance $F_\text{S}$. 
This is largely due to the significant improvement on $\text{Acc}_\text{S}^{(\mathcal{E}_\text{R})}$, indicating the predicted shading images refrain from changing at edges caused mainly by reflectance ($\mathcal{E}_\text{R}$). This is consistent with the qualitative analysis. Such manner for shading images is quite beneficial. 
With gradient constraint loss on top, the model shows comprehensive improvement, especially for reflectance. 
This agrees with our expectation. When shading is changing abruptly, reflectance is largely penalized and thus improve the $\text{Acc}_\text{R}^{(\mathcal{E}_\text{S})}$. When shading is smoothly changing, it also slightly penalize reflectance changes and thus improve $\text{RegionError}_\text{R}$. 
 % for shading and increase the reflectance edge performance $F_\text{R}$ and region errors. 
% This shows the gradient constraint prior in a self-supervised manner benefits the model. 
Comparing the results of CGIntrinsics and UNet-CLO, we can see that models trained with CloIntrinsics have much smaller shading region errors. 
This indicates models trained with CloIntrinsics generally fits tighter to the groundtruth, preserving more detailed information in shading. This validates the effectiveness of in-domain knowledge provided by diverse shapes and textures in CloIntrinsics training set. 
% The quantitative analysis is consistent with qualitative results shown in Fig.~\ref{fig:qualitative}. 
% This also validates the effectiveness of our proposed metrics. 

We also show cross-domain evaluation on MIT dataset in Table \ref{tbl:mit_quantitative}. 
Although the model is working on clothing domain, it still generalize well to MIT dataset. 
Our model predicts comparable reflectance and better shading compared to models trained with CGIntrinsics and BigTime. 

\vspace{1mm}
\noindent\textbf{Discussion. }
Our proposed diagnostic metrics provide more interpretable information on how model behaves, thus providing novel insights on this task. 
% The critical issue for intrinsic image decomposition 
The ill-posed nature of intrinsic image decomposition requires algorithms to disentangle local changes into different components. 
However, 
% in the literature of learning-based intrinsic image decomposition algorithms, 
the commonly employed regression and reconstruction loss tend to \emph{recall} more local changes and produce predictions with low \emph{precision}. 
To address this, we propose to use adversarial learning and a gradient exclusive loss to enforce both implicit and explicit reasoning on local changes. 
Similar adversarial module was employed in previous work \cite{lettry2018darn}, but we also stress the different insight for using such a module and the non-trivial effort on selecting effective data, i.e. the real-world shading set. 
The novel gradient constraint loss explicitly penalizes local entanglement and thus performs well on challenging clothing images. 

\subsection{Applications}
\label{sec:app}
% TODO!!!
\begin{figure}[t!]
\centering
\vspace{-1mm}
\includegraphics[width=0.95\linewidth]{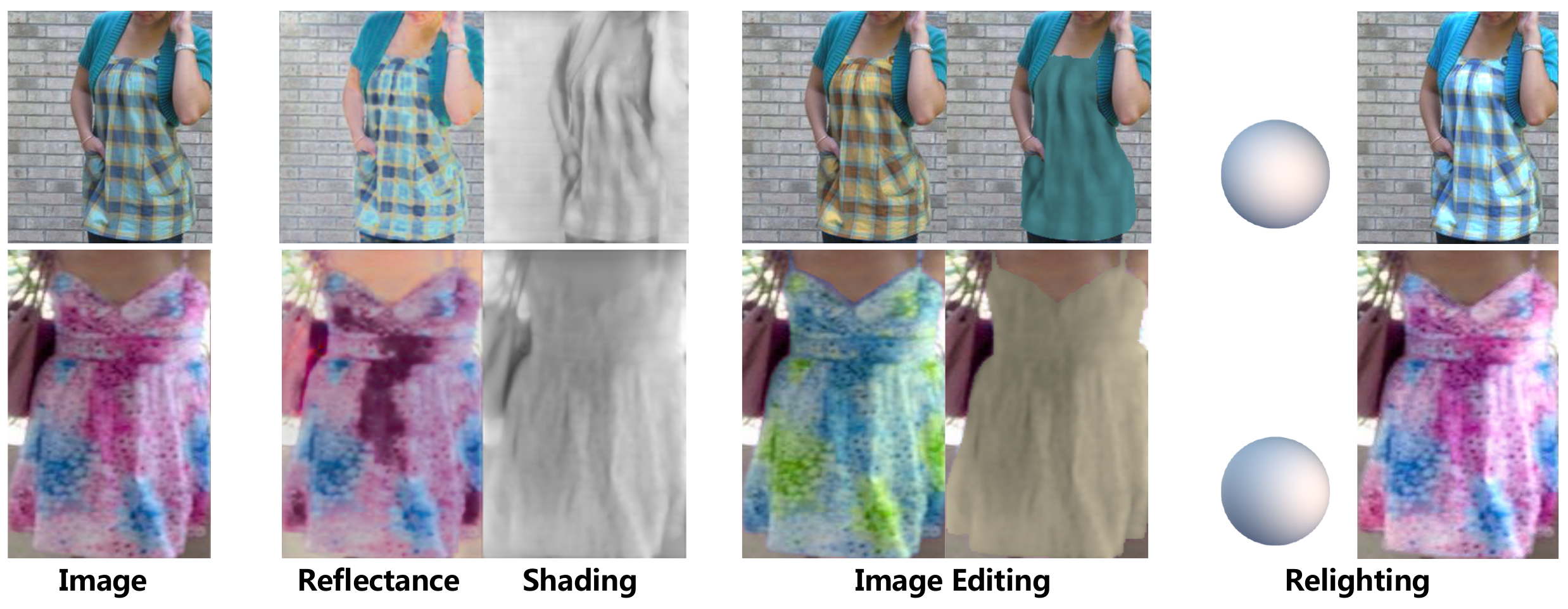} 
\vspace{-3mm}
\caption{\footnotesize Image editing and relighting with intrinsic images. 
% On these challenging cases, we perform albedo editing and relighting. Our ClothInNet model significantly reduces texture-copying artifacts and thus allows realistic editing. 
}
\label{fig:teaser}
\vspace{-3mm}
\end{figure}

With our intrinsic decomposition results of clothing images, we can perform graphics applications leveraging the embedded geometry information in the decomposed shading image and the color information in the reflectance image. 

\noindent\textbf{Color style editing.}
% As the clothing images can be recovered using the decomposed shading and reflectance, 
We fix the shading image and perform color style editing on the reflectance, as shown in Fig.~\ref{fig:teaser}. The image editing results show that the wrinkle and shape details are preserved. 
% and we can perform vivid results of the person wearing dress with another color style.

\noindent\textbf{Relighting.}
We iteratively optimize spherical harmonic lighting and normal map, following \cite{ChengleiShading}. 
The relighting result is shown in Fig.~\ref{fig:teaser}.

\begin{figure}[t!]
\centering
\vspace{-1mm}
\includegraphics[width=0.95\linewidth]{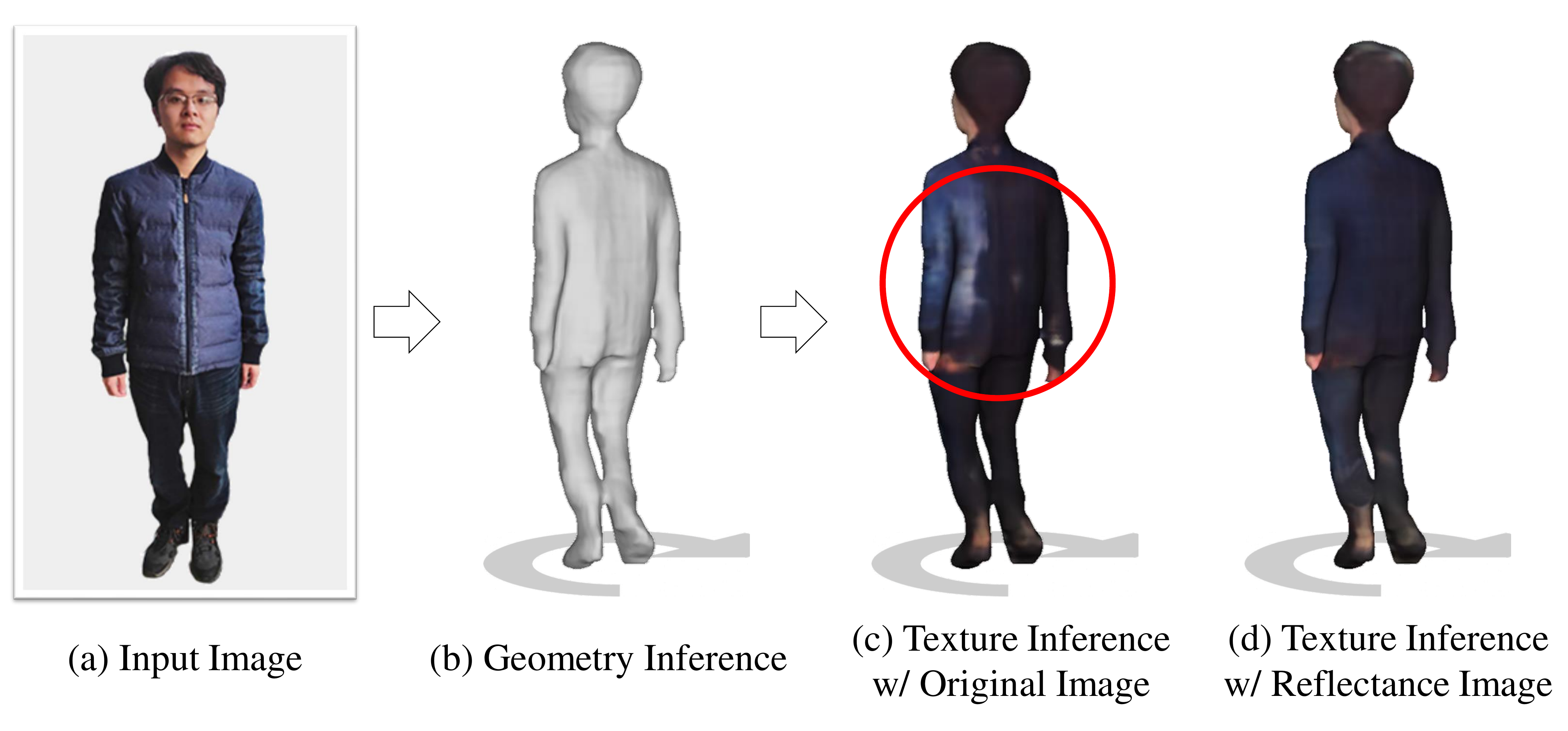} 
\vspace{-3mm}
\caption{\footnotesize Clothing geometry and texture inference using PIFU\cite{pifuSHNMKL19}.
% : (a) The input image, (b) the geometry inference result  using the original image as input, (c) the texture inference results using the original image as input and (d) the texture inference results using the reflectance image as input. 
}
\label{fig:texture_inference_app}
\vspace{-6mm}
\end{figure}

\noindent\textbf{Texture Inference.}
Due to the variation of lighting environments, inferencing full-body texture accurately remains challenging for single image human reconstruction models \cite{Zheng2019DeepHuman,SiCloPe2019,pifuSHNMKL19,tex2shape2019}. Our predicted reflectance image can be used for full-body texture inference. 
Specifically, given an in-the-wild image of a human, we can obtain the corresponding geometry and texture reconstruction result using PIFU\cite{pifuSHNMKL19}, as shown in Fig.\ref{fig:texture_inference_app}(a)-(c). However, as we can see in Fig.\ref{fig:texture_inference_app}(c), the texture inference  based on the original input image generates severe artifacts on the back side. In contrast, by feeding the reflectance image provided by our method, we can obtain a more plausible texture inference result in Fig.\ref{fig:texture_inference_app}(d). 

% Geometry inference from a single RGB image has been well studied in recent works like \cite{Zheng2019DeepHuman,SiCloPe2019,pifuSHNMKL19,tex2shape2019}, while inferencing full-body texture accurately remains challenging due to the variations of lighting environments. The color information in the reflectance image obtained using our method can be used for full-body texture inference. Specifically, taking an in-the-wild image of a human, we can obtain the corresponding geometry and texture reconstruction result using PIFU\cite{pifuSHNMKL19}, as shown in in Fig.\ref{fig:texture_inference_app}(a)-(c). However, as we can see in Fig.\ref{fig:texture_inference_app}(c), the texture inference  based on the original input image generates severe artifacts on the back side. In contrast, by feeding the reflectance image provided by our method, we can obtain a more plausible texture inference result. 

%In Fig.\ref{{fig:texture_inference_app}}, we show that given an in-the-wild image of a human and the corresponding geometry reconstruction results provided by PIFU\cite{pifuSHNMKL19}, the texture inference based on the original input image generates severe artifacts on the back side. In contrast, by taking  

%!TEX root=main.tex
% \vspace{-5mm}

\vspace{-2mm}
\section{Conclusion}
\label{sec:conc}
\vspace{-2mm}

In this paper, we extensively studied the task of intrinsic image decomposition for clothing images. 
We proposed a clothing intrinsic images dataset, an interpretable diagnostic evaluation metric, and a simple but effective model. 
% The synthetic training set helps produces sharp intrinsic images and 
The proposed adversarial method and carefully designed loss terms significantly reduces texture-copying artifacts, outperforming existing state-of-the-art approaches. 
The methods proposed in this paper are also able to extend to other domain, highlighting its value for intrinsic image decomposition research.

% use section* for acknowledgment
\ifCLASSOPTIONcompsoc
  % The Computer Society usually uses the plural form
  \section*{Acknowledgments}
\else
  % regular IEEE prefers the singular form
  \section*{Acknowledgment}
\fi

The authors would like to thank...

% Can use something like this to put references on a page
% by themselves when using endfloat and the captionsoff option.
\ifCLASSOPTIONcaptionsoff
  \newpage
\fi

% trigger a \newpage just before the given reference
% number - used to balance the columns on the last page
% adjust value as needed - may need to be readjusted if
% the document is modified later
%\IEEEtriggeratref{8}
% The "triggered" command can be changed if desired:
%\IEEEtriggercmd{\enlargethispage{-5in}}

% references section

% can use a bibliography generated by BibTeX as a .bbl file
% BibTeX documentation can be easily obtained at:
% http://mirror.ctan.org/biblio/bibtex/contrib/doc/
% The IEEEtran BibTeX style support page is at:
% http://www.michaelshell.org/tex/ieeetran/bibtex/
\bibliographystyle{IEEEtran}
% argument is your BibTeX string definitions and bibliography database(s)
\bibliography{egbib}

% Generated by IEEEtran.bst, version: 1.14 (2015/08/26)
\begin{thebibliography}{10}
\providecommand{\url}[1]{#1}
\csname url@samestyle\endcsname
\providecommand{\newblock}{\relax}
\providecommand{\bibinfo}[2]{#2}
\providecommand{\BIBentrySTDinterwordspacing}{\spaceskip=0pt\relax}
\providecommand{\BIBentryALTinterwordstretchfactor}{4}
\providecommand{\BIBentryALTinterwordspacing}{\spaceskip=\fontdimen2\font plus
\BIBentryALTinterwordstretchfactor\fontdimen3\font minus
  \fontdimen4\font\relax}
\providecommand{\BIBforeignlanguage}[2]{{%
\expandafter\ifx\csname l@#1\endcsname\relax
\typeout{** WARNING: IEEEtran.bst: No hyphenation pattern has been}%
\typeout{** loaded for the language `#1'. Using the pattern for}%
\typeout{** the default language instead.}%
\else
\language=\csname l@#1\endcsname
\fi
#2}}
\providecommand{\BIBdecl}{\relax}
\BIBdecl

\bibitem{imber2014intrinsic}
J.~Imber, J.-Y. Guillemaut, and A.~Hilton, ``Intrinsic textures for relightable
  free-viewpoint video,'' in \emph{European Conference on Computer
  Vision}.\hskip 1em plus 0.5em minus 0.4em\relax Springer, 2014, pp. 392--407.

\bibitem{wu2011shading}
C.~Wu, K.~Varanasi, Y.~Liu, H.-P. Seidel, and C.~Theobalt, ``Shading-based
  dynamic shape refinement from multi-view video under general illumination,''
  in \emph{2011 International Conference on Computer Vision}.\hskip 1em plus
  0.5em minus 0.4em\relax IEEE, 2011, pp. 1108--1115.

\bibitem{guo2017real}
K.~Guo, F.~Xu, T.~Yu, X.~Liu, Q.~Dai, and Y.~Liu, ``Real-time geometry, albedo,
  and motion reconstruction using a single rgb-d camera,'' \emph{ACM
  Transactions on Graphics (TOG)}, vol.~36, no.~3, p.~32, 2017.

\bibitem{meka2016live}
A.~Meka, M.~Zollh{\"o}fer, C.~Richardt, and C.~Theobalt, ``Live intrinsic
  video,'' \emph{ACM Transactions on Graphics (TOG)}, vol.~35, no.~4, p. 109,
  2016.

\bibitem{li2013capturing}
G.~Li, C.~Wu, C.~Stoll, Y.~Liu, K.~Varanasi, Q.~Dai, and C.~Theobalt,
  ``Capturing relightable human performances under general uncontrolled
  illumination,'' in \emph{Computer Graphics Forum}, vol.~32, no. 2pt3.\hskip
  1em plus 0.5em minus 0.4em\relax Wiley Online Library, 2013, pp. 275--284.

\bibitem{wu2011high}
C.~Wu, B.~Wilburn, Y.~Matsushita, and C.~Theobalt, ``High-quality shape from
  multi-view stereo and shading under general illumination,'' in \emph{CVPR
  2011}.\hskip 1em plus 0.5em minus 0.4em\relax IEEE, 2011, pp. 969--976.

\bibitem{land1971lightness}
\BIBentryALTinterwordspacing
E.~H. Land and J.~J. McCann, ``Lightness and retinex theory,'' \emph{J. Opt.
  Soc. Am.}, vol.~61, no.~1, pp. 1--11, Jan 1971. [Online]. Available:
  \url{http://www.osapublishing.org/abstract.cfm?URI=josa-61-1-1}
\BIBentrySTDinterwordspacing

\bibitem{omer2004color}
I.~Omer and M.~Werman, ``Color lines: Image specific color representation,'' in
  \emph{Proceedings of the 2004 IEEE Computer Society Conference on Computer
  Vision and Pattern Recognition, 2004. CVPR 2004.}, vol.~2.\hskip 1em plus
  0.5em minus 0.4em\relax IEEE, 2004, pp. II--II.

\bibitem{bousseau2009user}
A.~Bousseau, S.~Paris, and F.~Durand, ``User-assisted intrinsic images,'' in
  \emph{ACM Transactions on Graphics (TOG)}, vol.~28, no.~5.\hskip 1em plus
  0.5em minus 0.4em\relax ACM, 2009, p. 130.

\bibitem{rother2011recovering}
C.~Rother, M.~Kiefel, L.~Zhang, B.~Sch{\"o}lkopf, and P.~V. Gehler,
  ``Recovering intrinsic images with a global sparsity prior on reflectance,''
  in \emph{Advances in neural information processing systems}, 2011, pp.
  765--773.

\bibitem{garces2012intrinsic}
E.~Garces, A.~Munoz, J.~Lopez-Moreno, and D.~Gutierrez, ``Intrinsic images by
  clustering,'' in \emph{Computer graphics forum}, vol.~31, no.~4.\hskip 1em
  plus 0.5em minus 0.4em\relax Wiley Online Library, 2012, pp. 1415--1424.

\bibitem{zhao2012closed}
Q.~Zhao, P.~Tan, Q.~Dai, L.~Shen, E.~Wu, and S.~Lin, ``A closed-form solution
  to retinex with nonlocal texture constraints,'' \emph{IEEE transactions on
  pattern analysis and machine intelligence}, vol.~34, no.~7, pp. 1437--1444,
  2012.

\bibitem{liao2013non}
Z.~Liao, J.~Rock, Y.~Wang, and D.~Forsyth, ``Non-parametric filtering for
  geometric detail extraction and material representation,'' in
  \emph{Proceedings of the IEEE Conference on Computer Vision and Pattern
  Recognition}, 2013, pp. 963--970.

\bibitem{narihira2015direct}
T.~Narihira, M.~Maire, and S.~X. Yu, ``Direct intrinsics: Learning
  albedo-shading decomposition by convolutional regression,'' in
  \emph{Proceedings of the IEEE international conference on computer vision},
  2015, pp. 2992--2992.

\bibitem{shi2017learning}
J.~Shi, Y.~Dong, H.~Su, and S.~X. Yu, ``Learning non-lambertian object
  intrinsics across shapenet categories,'' in \emph{Proceedings of the IEEE
  Conference on Computer Vision and Pattern Recognition}, 2017, pp. 1685--1694.

\bibitem{nestmeyer2017reflectance}
T.~Nestmeyer and P.~V. Gehler, ``Reflectance adaptive filtering improves
  intrinsic image estimation,'' in \emph{Proceedings of the IEEE Conference on
  Computer Vision and Pattern Recognition}, 2017, pp. 6789--6798.

\bibitem{baslamisli2018cnn}
A.~S. Baslamisli, H.-A. Le, and T.~Gevers, ``Cnn based learning using
  reflection and retinex models for intrinsic image decomposition,'' in
  \emph{Proceedings of the IEEE Conference on Computer Vision and Pattern
  Recognition}, 2018, pp. 6674--6683.

\bibitem{fan2018revisiting}
Q.~Fan, J.~Yang, G.~Hua, B.~Chen, and D.~Wipf, ``Revisiting deep intrinsic
  image decompositions,'' in \emph{Proceedings of the IEEE conference on
  computer vision and pattern recognition}, 2018, pp. 8944--8952.

\bibitem{cheng2018intrinsic}
L.~Cheng, C.~Zhang, and Z.~Liao, ``Intrinsic image transformation via scale
  space decomposition,'' in \emph{Proceedings of the IEEE Conference on
  Computer Vision and Pattern Recognition}, 2018, pp. 656--665.

\bibitem{li2018cgintrinsics}
Z.~Li and N.~Snavely, ``Cgintrinsics: Better intrinsic image decomposition
  through physically-based rendering,'' in \emph{Proceedings of the European
  Conference on Computer Vision (ECCV)}, 2018, pp. 371--387.

\bibitem{butler2012naturalistic}
D.~J. Butler, J.~Wulff, G.~B. Stanley, and M.~J. Black, ``A naturalistic open
  source movie for optical flow evaluation,'' in \emph{European conference on
  computer vision}.\hskip 1em plus 0.5em minus 0.4em\relax Springer, 2012, pp.
  611--625.

\bibitem{baslamisli2018joint}
A.~S. Baslamisli, T.~T. Groenestege, P.~Das, H.-A. Le, S.~Karaoglu, and
  T.~Gevers, ``Joint learning of intrinsic images and semantic segmentation,''
  in \emph{Proceedings of the European Conference on Computer Vision (ECCV)},
  2018, pp. 286--302.

\bibitem{bell2014intrinsic}
S.~Bell, K.~Bala, and N.~Snavely, ``Intrinsic images in the wild,'' \emph{ACM
  Transactions on Graphics (TOG)}, vol.~33, no.~4, p. 159, 2014.

\bibitem{kovacs2017shading}
B.~Kovacs, S.~Bell, N.~Snavely, and K.~Bala, ``Shading annotations in the
  wild,'' in \emph{Proceedings of the IEEE Conference on Computer Vision and
  Pattern Recognition}, 2017, pp. 6998--7007.

\bibitem{janner2017self}
M.~Janner, J.~Wu, T.~D. Kulkarni, I.~Yildirim, and J.~Tenenbaum,
  ``Self-supervised intrinsic image decomposition,'' in \emph{Advances in
  Neural Information Processing Systems}, 2017, pp. 5936--5946.

\bibitem{li2018learning}
Z.~Li and N.~Snavely, ``Learning intrinsic image decomposition from watching
  the world,'' in \emph{Proceedings of the IEEE Conference on Computer Vision
  and Pattern Recognition}, 2018, pp. 9039--9048.

\bibitem{ma2018single}
W.-C. Ma, H.~Chu, B.~Zhou, R.~Urtasun, and A.~Torralba, ``Single image
  intrinsic decomposition without a single intrinsic image,'' in
  \emph{Proceedings of the European Conference on Computer Vision (ECCV)},
  2018, pp. 201--217.

\bibitem{grosse2009ground}
R.~Grosse, M.~K. Johnson, E.~H. Adelson, and W.~T. Freeman, ``Ground truth
  dataset and baseline evaluations for intrinsic image algorithms,'' in
  \emph{2009 IEEE 12th International Conference on Computer Vision}.\hskip 1em
  plus 0.5em minus 0.4em\relax IEEE, 2009, pp. 2335--2342.

\bibitem{Cloth3D}
\url{https://www.clo3d.com/}.

\bibitem{shen2008intrinsic}
L.~Shen, P.~Tan, and S.~Lin, ``Intrinsic image decomposition with non-local
  texture cues,'' in \emph{2008 IEEE Conference on Computer Vision and Pattern
  Recognition}.\hskip 1em plus 0.5em minus 0.4em\relax IEEE, 2008, pp. 1--7.

\bibitem{barron2014shape}
J.~T. Barron and J.~Malik, ``Shape, illumination, and reflectance from
  shading,'' \emph{IEEE transactions on pattern analysis and machine
  intelligence}, vol.~37, no.~8, pp. 1670--1687, 2014.

\bibitem{liu2020unsupervised}
Y.~Liu, Y.~Li, S.~You, and F.~Lu, ``Unsupervised learning for intrinsic image
  decomposition from a single image,'' in \emph{Proceedings of the IEEE/CVF
  Conference on Computer Vision and Pattern Recognition}, 2020, pp. 3248--3257.

\bibitem{zhu2020learning}
X.~Zhu, X.~Han, W.~Zhang, J.~Zhao, and L.~Liu, ``Learning intrinsic
  decomposition of complex-textured fashion images,'' in \emph{2020 IEEE
  International Conference on Multimedia and Expo (ICME)}.\hskip 1em plus 0.5em
  minus 0.4em\relax IEEE, 2020, pp. 1--6.

\bibitem{li2021sparse}
K.~Li, Y.~Wang, X.~Ye, C.~Yan, and J.~Yang, ``Sparse intrinsic decomposition
  and applications,'' \emph{Signal Processing: Image Communication}, vol.~95,
  p. 116281, 2021.

\bibitem{lettry2018darn}
L.~Lettry, K.~Vanhoey, and L.~Van~Gool, ``Darn: a deep adversarial residual
  network for intrinsic image decomposition,'' in \emph{2018 IEEE Winter
  Conference on Applications of Computer Vision (WACV)}.\hskip 1em plus 0.5em
  minus 0.4em\relax IEEE, 2018, pp. 1359--1367.

\bibitem{sengupta2019neural}
S.~Sengupta, J.~Gu, K.~Kim, G.~Liu, D.~W. Jacobs, and J.~Kautz, ``Neural
  inverse rendering of an indoor scene from a single image,'' \emph{arXiv
  preprint arXiv:1901.02453}, 2019.

\bibitem{yu2019inverserendernet}
Y.~Yu and W.~A. Smith, ``Inverserendernet: Learning single image inverse
  rendering,'' in \emph{Proceedings of the IEEE Conference on Computer Vision
  and Pattern Recognition}, 2019, pp. 3155--3164.

\bibitem{duchene2015multi}
S.~Duch{\^e}ne, C.~Riant, G.~Chaurasia, J.~Lopez-Moreno, P.-Y. Laffont,
  S.~Popov, A.~Bousseau, and G.~Drettakis, ``Multi-view intrinsic images of
  outdoors scenes with an application to relighting,'' \emph{ACM Transactions
  on Graphics (TOG)}, 2015.

\bibitem{lettry2018unsupervised}
L.~Lettry, K.~Vanhoey, and L.~Van~Gool, ``Unsupervised deep single-image
  intrinsic decomposition using illumination-varying image sequences,'' in
  \emph{Computer Graphics Forum}, vol.~37, no.~7.\hskip 1em plus 0.5em minus
  0.4em\relax Wiley Online Library, 2018, pp. 409--419.

\bibitem{bi2018deep}
S.~Bi, N.~K. Kalantari, and R.~Ramamoorthi, ``Deep hybrid real and synthetic
  training for intrinsic decomposition,'' 2018.

\bibitem{chang2015shapenet}
A.~X. Chang, T.~Funkhouser, L.~Guibas, P.~Hanrahan, Q.~Huang, Z.~Li,
  S.~Savarese, M.~Savva, S.~Song, H.~Su \emph{et~al.}, ``Shapenet: An
  information-rich 3d model repository,'' \emph{arXiv preprint
  arXiv:1512.03012}, 2015.

\bibitem{Blender}
\url{https://www.blender.org/}.

\bibitem{yamaguchi2013paper}
K.~Yamaguchi, M.~Hadi~Kiapour, and T.~L. Berg, ``Paper doll parsing: Retrieving
  similar styles to parse clothing items,'' in \emph{Proceedings of the IEEE
  international conference on computer vision}, 2013, pp. 3519--3526.

\bibitem{canny1986computational}
J.~Canny, ``A computational approach to edge detection,'' \emph{IEEE
  Transactions on pattern analysis and machine intelligence}, no.~6, pp.
  679--698, 1986.

\bibitem{ATR}
X.~Liang, S.~Liu, X.~Shen, J.~Yang, L.~Liu, J.~Dong, L.~Lin, and S.~Yan, ``Deep
  human parsing with active template regression,'' \emph{IEEE Transactions on
  Pattern Analysis and Machine Intelligence}, vol.~37, no.~12, pp. 2402--2414,
  2015.

\bibitem{CO-CNN}
X.~Liang, C.~Xu, X.~Shen, J.~Yang, S.~Liu, J.~Tang, L.~Lin, and S.~Yan, ``Human
  parsing with contextualized convolutional neural network,'' \emph{IEEE
  Transactions on Pattern Analysis and Machine Intelligence}, vol.~39, no.~1,
  pp. 115--127, Jan 2017.

\bibitem{land1971lightness1}
E.~H. Land and J.~J. McCann, ``Lightness and retinex theory,'' \emph{Josa},
  vol.~61, no.~1, pp. 1--11, 1971.

\bibitem{ronneberger2015u}
O.~Ronneberger, P.~Fischer, and T.~Brox, ``U-net: Convolutional networks for
  biomedical image segmentation,'' in \emph{International Conference on Medical
  image computing and computer-assisted intervention}.\hskip 1em plus 0.5em
  minus 0.4em\relax Springer, 2015, pp. 234--241.

\bibitem{he2016deep}
K.~He, X.~Zhang, S.~Ren, and J.~Sun, ``Deep residual learning for image
  recognition,'' in \emph{Proceedings of the IEEE conference on computer vision
  and pattern recognition}, 2016, pp. 770--778.

\bibitem{ioffe2015batch}
S.~Ioffe and C.~Szegedy, ``Batch normalization: Accelerating deep network
  training by reducing internal covariate shift,'' \emph{arXiv preprint
  arXiv:1502.03167}, 2015.

\bibitem{glorot2011deep}
X.~Glorot, A.~Bordes, and Y.~Bengio, ``Deep sparse rectifier neural networks,''
  in \emph{Proceedings of the fourteenth international conference on artificial
  intelligence and statistics}, 2011, pp. 315--323.

\bibitem{ChengleiShading}
C.~Wu, B.~Wilburn, Y.~Matsushita, and C.~Theobalt, ``High-quality shape from
  multi-view stereo and shading under general illumination,'' in \emph{CVPR},
  2011, pp. 969--976.

\bibitem{pifuSHNMKL19}
S.~Saito, , Z.~Huang, R.~Natsume, S.~Morishima, A.~Kanazawa, and H.~Li, ``Pifu:
  Pixel-aligned implicit function for high-resolution clothed human
  digitization,'' in \emph{{IEEE} International Conference on Computer Vision
  ({ICCV})}, 2019.

\bibitem{Zheng2019DeepHuman}
Z.~Zheng, T.~Yu, Y.~Wei, Q.~Dai, and Y.~Liu, ``Deephuman: 3d human
  reconstruction from a single image,'' in \emph{{IEEE} International
  Conference on Computer Vision ({ICCV})}, 2019.

\bibitem{SiCloPe2019}
\BIBentryALTinterwordspacing
R.~Natsume, S.~Saito, Z.~Huang, W.~Chen, C.~Ma, H.~Li, and S.~Morishima,
  ``Siclope: Silhouette-based clothed people,'' \emph{CoRR}, vol.
  abs/1901.00049, 2019. [Online]. Available:
  \url{http://arxiv.org/abs/1901.00049}
\BIBentrySTDinterwordspacing

\bibitem{tex2shape2019}
T.~Alldieck, G.~Pons-Moll, C.~Theobalt, and M.~Magnor, ``Tex2shape: Detailed
  full human body geometry from a single image,'' in \emph{{IEEE} International
  Conference on Computer Vision ({ICCV})}, 2019.

\end{thebibliography}
\end{document}

% \renewcommand\thelinenumber{\color[rgb]{0.2,0.5,0.8}\normalfont\sffamily\scriptsize\arabic{linenumber}\color[rgb]{0,0,0}}
% \renewcommand\makeLineNumber {\hss\thelinenumber\ \hspace{6mm} \rlap{\hskip\textwidth\ \hspace{6.5mm}\thelinenumber}}
% \linenumbers
\pagestyle{headings}
\mainmatter
\def\ECCVSubNumber{3319}  % Insert your submission number here

%%%%%%%%% TITLE
\title{Supplementary Material:\\ Learning Intrinsic Images for Clothing}

\titlerunning{ECCV-20 submission ID \ECCVSubNumber} 
\authorrunning{ECCV-20 submission ID \ECCVSubNumber} 
\author{Anonymous ECCV submission}
\institute{Paper ID \ECCVSubNumber}
% \author{First Author\\
% Institution1\\
% Institution1 address\\
% {\tt\small firstauthor@i1.org}
% % For a paper whose authors are all at the same institution,
% % omit the following lines up until the closing ``}''.
% % Additional authors and addresses can be added with ``\and'',
% % just like the second author.
% % To save space, use either the email address or home page, not both
% \and
% Second Author\\
% Institution2\\
% First line of institution2 address\\
% {\tt\small secondauthor@i2.org}
% }

\maketitle
%\thispagestyle{empty}

%%%%%%%%% BODY TEXT

\vspace{3mm}
In this supplementary material, we first provide a more detailed discussion on common artifacts produced by current state-of-the-art methods. Then we describe the motivation and intuition on the proposed edge-aware metric and the details of data annotation. We also present more qualitative results of ClothInNet model on real world clothing images. 

\vspace{3mm}
\noindent\textbf{Common Artifacts of Intrinsic Image Decomposition.} 
To evaluate and improve intrinsic image decomposition methods, it's important to understand common artifacts produced by intrinsic image models. 
It's pretty easy for an expert with knowledge of intrinsic images to discriminate the predicted results, by looking at the artifacts around edges and regions of interest. 

\begin{figure}[h]
\begin{center}
\scalebox{1.0}[0.6]{\includegraphics[height=5cm]{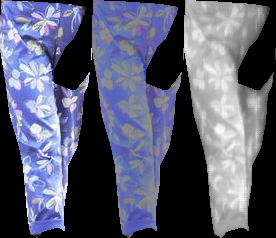}} 
\scalebox{1.0}[1.0]{\includegraphics[height=3cm]{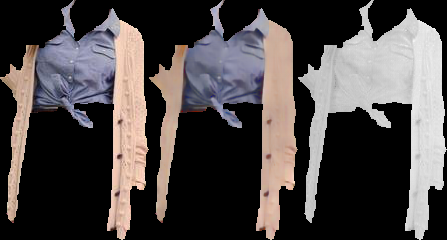}} \\
\end{center}
\vspace{-4mm}
\caption{Common artifacts of state-of-the-art methods, texture-copying (left) and region error (right). From left to right are  RGB image, reflectance and shading. }
\label{fig:common}
\vspace{-4mm}
\end{figure}

Edges $\mathcal{E}$ reflect changes of RGB image, and can be caused by reflectance only $\mathcal{E}_\text{R}$, shading only $\mathcal{E}_\text{S}$ or both $\mathcal{E}_\text{RS}$. 
An expert can easily discriminate the predicted results with $\mathcal{E}_\text{R}$ and $\mathcal{E}_\text{S}$. 
The instance in the left of Fig.~\ref{fig:common} shows an example of common texture-copying artifacts around edges. 
The wrinkles caused by shape deformation should not be included in the predicted reflectance image. The shading image on the right should also not include the flower texture. 
%TODO: performance around $\mathcal{E}_\text{R}$ and $\mathcal{E}_\text{S}$ are important. 

Apart from edge performance, the predicted intrinsic images should also behave well in regions of the whole image. 
Human can easily identify constant-reflectance regions. 
In these regions, reflectance shouldn't show large changes and shading should account for most of these changes and fit tightly to the local deformation. 
The example in the right of Fig.~\ref{fig:common} shows the region error. In the purple shirt region, reflectance should not vary , and shading should change following the wrinkle pattern. 

These challenging artifacts are critical for intrinsic image decomposition algorithms and deserve special attention.

\vspace{3mm}
\noindent\textbf{Edge-Aware Metric.} 
IIW dataset \cite{bell2014intrinsic} and SAW dataset \cite{kovacs2017shading} are popular real-world benchmark datasets, but the metrics can hardly evaluate the common artifacts mentioned above. 
IIW dataset uses weighted human disagreement rate (WHDR) to evaluate how the reflectance prediction agrees with sparse annotation. 
It can detect the region error of reflectance but fails to indicate edge performance. 
SAW dataset includes the evaluation of predicted shading images around $\mathcal{E}_\text{S}$, but still cannot evaluate the performance around $\mathcal{E}_\text{R}$. Also, SAW dataset doesn't provide region errors and can't indicate how well the predicted shading fits to the local changes. 

Motivated by the above discussion, we propose to let human annotate both edges and regions of interest. We annotate constant-reflectance regions ($\mathcal{R}$) and edges caused by reflectance only ($\mathcal{E}_\text{R}$) and shading only ($\mathcal{E}_\text{S}$). 
For constant-reflectance regions, we define $\text{RegionError}_\text{R}$ and $\text{RegionError}_\text{S}$ as metrics, which show the variance of reflectance and scale invariant MSE of shading within these regions. 
For edges, we use $\text{Acc}_\text{R}^{(\mathcal{E}_\text{S})}, \text{Acc}_\text{R}^{(\mathcal{E}_\text{R})}, \text{Acc}_\text{S}^{(\mathcal{E}_\text{R})}$ and $ \text{Acc}_\text{S}^{(\mathcal{E}_\text{S})}$ to give diagnostic information on the component entanglement around edges. 
$\text{Acc}_\text{R}^{(\mathcal{E}_\text{R})}$ and $\text{Acc}_\text{S}^{(\mathcal{E}_\text{S})}$ are an approximation of recall, and $\text{Acc}_\text{R}^{(\mathcal{E}_\text{S})}$ and $\text{Acc}_\text{S}^{(\mathcal{E}_\text{R})}$ are positively correlated to precision. These metrics shows how models perform around edges. 
We finally use a weighted harmonic mean $F_\text{R}$ and $F_\text{S}$ to indicate general edge performance of reflectance and shading respectively. 

\vspace{3mm}
\noindent\textbf{Real-World Testing Set Annotation.}
The annotation collection pipeline includes data annotation and data verification, conducted by three annotators with background in intrinsic images. 
For data annotation, we ask two of the annotators to annotate the images twice. They are asked to only annotate the confident parts. 
The annotation will then be combined, and another inspector will verify the annotation. 
We noticed that it's hard for the inspector to find mistakes directly by watching the edge annotation, but it's easier for the inspector to judge whether reflectance and shading are well behaved. 
Based on this observation, we incorporate an interactive method \cite{bousseau2009user} for annotation inspection. 
Specifically, We use our edge annotation as the prior to solve for intrinsic images, and the results will be displayed to the annotator. 
Only when our annotation leads to plausible decomposition results can we assume the annotation is reliable. 
% After considering R2's constructive comments, we ask three more annotators to annotate the edges and use a voting strategy to only pick out confident annotation before verification. This further improves edge robustness. 
% We believe the professional annotators and the strategy are enough for good quality annotation. 
With the above professional annotation procedure, we obtained high quality annotation on real-world images.

\vspace{3mm}
\noindent\textbf{ClothInNet Model.} 
To address the common artifacts mentioned above, we propose ClothInNet model. 
The adversarial module serves the role as a human expert mentioned above, and significantly improves the edge performance of shading and removes the texture patterns in shading. Here we stress the importance of our collected in-domain data, i.e. the real-world shading set. It thus enforce the network predictions to match with the real-world shading distribution. 
The gradient constraint loss introduces explicit reasoning around edges, further improving the results, especially the performance of reflectance. Compare to prior state-of-the-art methods, it significantly removes the wrinkle patterns in reflectance. 
% We also show the proposed CloIntrinsics training set helps improve region performance in the ablation study. 
More qualitative results are shown in Fig.~\ref{fig:qualitative1},~\ref{fig:qualitative2},~\ref{fig:qualitative3},~\ref{fig:qualitative4}. 

% \begin{figure*}[t]
% \begin{center}
% \scalebox{1.0}[1.0]{\includegraphics[width=0.8\textwidth]{src/supp1.pdf}} \\
% \end{center}
% \caption{Qualitative Results of ClothInNet. From left to right are input image, predicted reflectance and shading image. }
% \label{fig:supp1}
% \vspace{-2mm}
% \end{figure*}

% \begin{figure*}[t]
% \begin{center}
% \scalebox{1.0}[1.0]{\includegraphics[width=0.8\textwidth]{src/supp3.pdf}} \\
% \end{center}
% \vspace{-2mm}
% \caption{Qualitative Results of ClothInNet. From left to right are input image, predicted reflectance and shading image. }
% \label{fig:supp2}
% \vspace{-2mm}
% \end{figure*}

% \begin{figure*}[t]
% \begin{center}
% \scalebox{1.0}[1.0]{\includegraphics[width=0.8\textwidth]{src/supp2.pdf}} \\
% \end{center}
% \caption{Qualitative Results of ClothInNet. From left to right are input image, predicted reflectance and shading image. }
% \label{fig:supp3}
% \vspace{-2mm}
% \end{figure*}

\begin{figure*}[t!]
\centering
\vspace{-3mm}
\begin{tabular}{@{}c@{}c@{}c@{}c@{}c@{}c@{}}
\scalebox{1.0}[1.0]{\includegraphics[width=0.165\textwidth]{src/supp/I_00008.png}} & 
\scalebox{1.0}[1.0]{\includegraphics[width=0.165\textwidth]{src/supp/R_Retinex_00008.png}} & 
\scalebox{1.0}[1.0]{\includegraphics[width=0.165\textwidth]{src/supp/R_Zhao_00008.png}} & 
\scalebox{1.0}[1.0]{\includegraphics[width=0.165\textwidth]{src/supp/R_BT_00008.png}} & 
\scalebox{1.0}[1.0]{\includegraphics[width=0.165\textwidth]{src/supp/R_CGI_00008.png}} & 
\scalebox{1.0}[1.0]{\includegraphics[width=0.165\textwidth]{src/supp/R_CLO_00008.png}} \\ & 
\scalebox{1.0}[1.0]{\includegraphics[width=0.165\textwidth]{src/supp/S_Retinex_00008.png}} & 
\scalebox{1.0}[1.0]{\includegraphics[width=0.165\textwidth]{src/supp/S_Zhao_00008.png}} & 
\scalebox{1.0}[1.0]{\includegraphics[width=0.165\textwidth]{src/supp/S_BT_00008.png}} & 
\scalebox{1.0}[1.0]{\includegraphics[width=0.165\textwidth]{src/supp/S_CGI_00008.png}} & 
\scalebox{1.0}[1.0]{\includegraphics[width=0.165\textwidth]{src/supp/S_CLO_00008.png}} \\

\scalebox{1.0}[1.0]{\includegraphics[width=0.165\textwidth]{src/supp/I_00005.png}} & 
\scalebox{1.0}[1.0]{\includegraphics[width=0.165\textwidth]{src/supp/R_Retinex_00005.png}} & 
\scalebox{1.0}[1.0]{\includegraphics[width=0.165\textwidth]{src/supp/R_Zhao_00005.png}} & 
\scalebox{1.0}[1.0]{\includegraphics[width=0.165\textwidth]{src/supp/R_BT_00005.png}} & 
\scalebox{1.0}[1.0]{\includegraphics[width=0.165\textwidth]{src/supp/R_CGI_00005.png}} & 
\scalebox{1.0}[1.0]{\includegraphics[width=0.165\textwidth]{src/supp/R_CLO_00005.png}} \\ & 
\scalebox{1.0}[1.0]{\includegraphics[width=0.165\textwidth]{src/supp/S_Retinex_00005.png}} & 
\scalebox{1.0}[1.0]{\includegraphics[width=0.165\textwidth]{src/supp/S_Zhao_00005.png}} & 
\scalebox{1.0}[1.0]{\includegraphics[width=0.165\textwidth]{src/supp/S_BT_00005.png}} & 
\scalebox{1.0}[1.0]{\includegraphics[width=0.165\textwidth]{src/supp/S_CGI_00005.png}} & 
\scalebox{1.0}[1.0]{\includegraphics[width=0.165\textwidth]{src/supp/S_CLO_00005.png}} \\

\scalebox{1.0}[1.0]{\includegraphics[width=0.165\textwidth]{src/supp/I_00001.png}} & 
\scalebox{1.0}[1.0]{\includegraphics[width=0.165\textwidth]{src/supp/R_Retinex_00001.png}} & 
\scalebox{1.0}[1.0]{\includegraphics[width=0.165\textwidth]{src/supp/R_Zhao_00001.png}} & 
\scalebox{1.0}[1.0]{\includegraphics[width=0.165\textwidth]{src/supp/R_BT_00001.png}} & 
\scalebox{1.0}[1.0]{\includegraphics[width=0.165\textwidth]{src/supp/R_CGI_00001.png}} & 
\scalebox{1.0}[1.0]{\includegraphics[width=0.165\textwidth]{src/supp/R_CLO_00001.png}} \\ & 
\scalebox{1.0}[1.0]{\includegraphics[width=0.165\textwidth]{src/supp/S_Retinex_00001.png}} & 
\scalebox{1.0}[1.0]{\includegraphics[width=0.165\textwidth]{src/supp/S_Zhao_00001.png}} & 
\scalebox{1.0}[1.0]{\includegraphics[width=0.165\textwidth]{src/supp/S_BT_00001.png}} & 
\scalebox{1.0}[1.0]{\includegraphics[width=0.165\textwidth]{src/supp/S_CGI_00001.png}} & 
\scalebox{1.0}[1.0]{\includegraphics[width=0.165\textwidth]{src/supp/S_CLO_00001.png}} \\

{\footnotesize Image } & 
{\footnotesize Retinex \cite{land1971lightness1} } & 
{\footnotesize Zhao et al. \cite{zhao2012closed} } & 
{\footnotesize BigTime \cite{li2018learning}} & 
{\footnotesize CGIntrinsics \cite{li2018cgintrinsics}} & 
{\footnotesize ClothInNet } \\
\end{tabular}
\vspace{-3mm}
\caption{ \textbf{Qualitative comparisons on real world clothing images. } 
For each method, the first row is reflectance and the second row is shading. 
Our model significantly removes texture-copying artifacts while retaining tiny details. 
The CGIntrinsics model is trained on CGI, IIW and SAW. 
}
\label{fig:qualitative1} 
\vspace{-3mm}
\end{figure*}

\begin{figure*}[t!]
\centering
\vspace{-3mm}
\begin{tabular}{@{}c@{}c@{}c@{}c@{}c@{}c@{}}
\scalebox{1.0}[1.0]{\includegraphics[width=0.165\textwidth]{src/supp/I_00004.png}} & 
\scalebox{1.0}[1.0]{\includegraphics[width=0.165\textwidth]{src/supp/R_Retinex_00004.png}} & 
\scalebox{1.0}[1.0]{\includegraphics[width=0.165\textwidth]{src/supp/R_Zhao_00004.png}} & 
\scalebox{1.0}[1.0]{\includegraphics[width=0.165\textwidth]{src/supp/R_BT_00004.png}} & 
\scalebox{1.0}[1.0]{\includegraphics[width=0.165\textwidth]{src/supp/R_CGI_00004.png}} & 
\scalebox{1.0}[1.0]{\includegraphics[width=0.165\textwidth]{src/supp/R_CLO_00004.png}} \\ & 
\scalebox{1.0}[1.0]{\includegraphics[width=0.165\textwidth]{src/supp/S_Retinex_00004.png}} & 
\scalebox{1.0}[1.0]{\includegraphics[width=0.165\textwidth]{src/supp/S_Zhao_00004.png}} & 
\scalebox{1.0}[1.0]{\includegraphics[width=0.165\textwidth]{src/supp/S_BT_00004.png}} & 
\scalebox{1.0}[1.0]{\includegraphics[width=0.165\textwidth]{src/supp/S_CGI_00004.png}} & 
\scalebox{1.0}[1.0]{\includegraphics[width=0.165\textwidth]{src/supp/S_CLO_00004.png}} \\

\scalebox{1.0}[0.8]{\includegraphics[width=0.165\textwidth]{src/supp/I_00000.png}} & 
\scalebox{1.0}[0.8]{\includegraphics[width=0.165\textwidth]{src/supp/R_Retinex_00000.png}} & 
\scalebox{1.0}[0.8]{\includegraphics[width=0.165\textwidth]{src/supp/R_Zhao_00000.png}} & 
\scalebox{1.0}[0.8]{\includegraphics[width=0.165\textwidth]{src/supp/R_BT_00000.png}} & 
\scalebox{1.0}[0.8]{\includegraphics[width=0.165\textwidth]{src/supp/R_CGI_00000.png}} & 
\scalebox{1.0}[0.8]{\includegraphics[width=0.165\textwidth]{src/supp/R_CLO_00000.png}} \\ & 
\scalebox{1.0}[0.8]{\includegraphics[width=0.165\textwidth]{src/supp/S_Retinex_00000.png}} & 
\scalebox{1.0}[0.8]{\includegraphics[width=0.165\textwidth]{src/supp/S_Zhao_00000.png}} & 
\scalebox{1.0}[0.8]{\includegraphics[width=0.165\textwidth]{src/supp/S_BT_00000.png}} & 
\scalebox{1.0}[0.8]{\includegraphics[width=0.165\textwidth]{src/supp/S_CGI_00000.png}} & 
\scalebox{1.0}[0.8]{\includegraphics[width=0.165\textwidth]{src/supp/S_CLO_00000.png}} \\

{\footnotesize Image } & 
{\footnotesize Retinex \cite{land1971lightness1} } & 
{\footnotesize Zhao et al. \cite{zhao2012closed} } & 
{\footnotesize BigTime \cite{li2018learning}} & 
{\footnotesize CGIntrinsics \cite{li2018cgintrinsics}} & 
{\footnotesize ClothInNet } \\
\end{tabular}
\vspace{-3mm}
\caption{ \textbf{Qualitative comparisons on real world clothing images. } 
For each method, the first row is reflectance and the second row is shading. 
Our model significantly removes texture-copying artifacts while retaining tiny details. 
The CGIntrinsics model is trained on CGI, IIW and SAW. 
}
\label{fig:qualitative2} 
\vspace{-3mm}
\end{figure*}

\begin{figure*}[t!]
\centering
\vspace{-3mm}
\begin{tabular}{@{}c@{}c@{}c@{}c@{}c@{}c@{}}
\scalebox{1.0}[0.9]{\includegraphics[width=0.165\textwidth]{src/supp/I_00003.png}} & 
\scalebox{1.0}[0.9]{\includegraphics[width=0.165\textwidth]{src/supp/R_Retinex_00003.png}} & 
\scalebox{1.0}[0.9]{\includegraphics[width=0.165\textwidth]{src/supp/R_Zhao_00003.png}} & 
\scalebox{1.0}[0.9]{\includegraphics[width=0.165\textwidth]{src/supp/R_BT_00003.png}} & 
\scalebox{1.0}[0.9]{\includegraphics[width=0.165\textwidth]{src/supp/R_CGI_00003.png}} & 
\scalebox{1.0}[0.9]{\includegraphics[width=0.165\textwidth]{src/supp/R_CLO_00003.png}} \\ & 
\scalebox{1.0}[0.9]{\includegraphics[width=0.165\textwidth]{src/supp/S_Retinex_00003.png}} & 
\scalebox{1.0}[0.9]{\includegraphics[width=0.165\textwidth]{src/supp/S_Zhao_00003.png}} & 
\scalebox{1.0}[0.9]{\includegraphics[width=0.165\textwidth]{src/supp/S_BT_00003.png}} & 
\scalebox{1.0}[0.9]{\includegraphics[width=0.165\textwidth]{src/supp/S_CGI_00003.png}} & 
\scalebox{1.0}[0.9]{\includegraphics[width=0.165\textwidth]{src/supp/S_CLO_00003.png}} \\

\scalebox{1.0}[0.9]{\includegraphics[width=0.165\textwidth]{src/supp/I_00007.png}} & 
\scalebox{1.0}[0.9]{\includegraphics[width=0.165\textwidth]{src/supp/R_Retinex_00007.png}} & 
\scalebox{1.0}[0.9]{\includegraphics[width=0.165\textwidth]{src/supp/R_Zhao_00007.png}} & 
\scalebox{1.0}[0.9]{\includegraphics[width=0.165\textwidth]{src/supp/R_BT_00007.png}} & 
\scalebox{1.0}[0.9]{\includegraphics[width=0.165\textwidth]{src/supp/R_CGI_00007.png}} & 
\scalebox{1.0}[0.9]{\includegraphics[width=0.165\textwidth]{src/supp/R_CLO_00007.png}} \\ & 
\scalebox{1.0}[0.9]{\includegraphics[width=0.165\textwidth]{src/supp/S_Retinex_00007.png}} & 
\scalebox{1.0}[0.9]{\includegraphics[width=0.165\textwidth]{src/supp/S_Zhao_00007.png}} & 
\scalebox{1.0}[0.9]{\includegraphics[width=0.165\textwidth]{src/supp/S_BT_00007.png}} & 
\scalebox{1.0}[0.9]{\includegraphics[width=0.165\textwidth]{src/supp/S_CGI_00007.png}} & 
\scalebox{1.0}[0.9]{\includegraphics[width=0.165\textwidth]{src/supp/S_CLO_00007.png}} \\

{\footnotesize Image } & 
{\footnotesize Retinex \cite{land1971lightness1} } & 
{\footnotesize Zhao et al. \cite{zhao2012closed} } & 
{\footnotesize BigTime \cite{li2018learning}} & 
{\footnotesize CGIntrinsics \cite{li2018cgintrinsics}} & 
{\footnotesize ClothInNet } \\
\end{tabular}
\vspace{-3mm}
\caption{ \textbf{Qualitative comparisons on real world clothing images. } 
For each method, the first row is reflectance and the second row is shading. 
Our model significantly removes texture-copying artifacts while retaining tiny details. 
The CGIntrinsics model is trained on CGI, IIW and SAW. 
}
\label{fig:qualitative3} 
\vspace{-3mm}
\end{figure*}

\begin{figure*}[t!]
\centering
\vspace{-3mm}
\begin{tabular}{@{}c@{}c@{}c@{}c@{}c@{}c@{}}
\scalebox{1.0}[1.0]{\includegraphics[width=0.165\textwidth]{src/supp/I_00002.png}} & 
\scalebox{1.0}[1.0]{\includegraphics[width=0.165\textwidth]{src/supp/R_Retinex_00002.png}} & 
\scalebox{1.0}[1.0]{\includegraphics[width=0.165\textwidth]{src/supp/R_Zhao_00002.png}} & 
\scalebox{1.0}[1.0]{\includegraphics[width=0.165\textwidth]{src/supp/R_BT_00002.png}} & 
\scalebox{1.0}[1.0]{\includegraphics[width=0.165\textwidth]{src/supp/R_CGI_00002.png}} & 
\scalebox{1.0}[1.0]{\includegraphics[width=0.165\textwidth]{src/supp/R_CLO_00002.png}} \\ & 
\scalebox{1.0}[1.0]{\includegraphics[width=0.165\textwidth]{src/supp/S_Retinex_00002.png}} & 
\scalebox{1.0}[1.0]{\includegraphics[width=0.165\textwidth]{src/supp/S_Zhao_00002.png}} & 
\scalebox{1.0}[1.0]{\includegraphics[width=0.165\textwidth]{src/supp/S_BT_00002.png}} & 
\scalebox{1.0}[1.0]{\includegraphics[width=0.165\textwidth]{src/supp/S_CGI_00002.png}} & 
\scalebox{1.0}[1.0]{\includegraphics[width=0.165\textwidth]{src/supp/S_CLO_00002.png}} \\

\scalebox{1.0}[1.0]{\includegraphics[width=0.165\textwidth]{src/supp/I_00006.png}} & 
\scalebox{1.0}[1.0]{\includegraphics[width=0.165\textwidth]{src/supp/R_Retinex_00006.png}} & 
\scalebox{1.0}[1.0]{\includegraphics[width=0.165\textwidth]{src/supp/R_Zhao_00006.png}} & 
\scalebox{1.0}[1.0]{\includegraphics[width=0.165\textwidth]{src/supp/R_BT_00006.png}} & 
\scalebox{1.0}[1.0]{\includegraphics[width=0.165\textwidth]{src/supp/R_CGI_00006.png}} & 
\scalebox{1.0}[1.0]{\includegraphics[width=0.165\textwidth]{src/supp/R_CLO_00006.png}} \\ & 
\scalebox{1.0}[1.0]{\includegraphics[width=0.165\textwidth]{src/supp/S_Retinex_00006.png}} & 
\scalebox{1.0}[1.0]{\includegraphics[width=0.165\textwidth]{src/supp/S_Zhao_00006.png}} & 
\scalebox{1.0}[1.0]{\includegraphics[width=0.165\textwidth]{src/supp/S_BT_00006.png}} & 
\scalebox{1.0}[1.0]{\includegraphics[width=0.165\textwidth]{src/supp/S_CGI_00006.png}} & 
\scalebox{1.0}[1.0]{\includegraphics[width=0.165\textwidth]{src/supp/S_CLO_00006.png}} \\

{\footnotesize Image } & 
{\footnotesize Retinex \cite{land1971lightness1} } & 
{\footnotesize Zhao et al. \cite{zhao2012closed} } & 
{\footnotesize BigTime \cite{li2018learning}} & 
{\footnotesize CGIntrinsics \cite{li2018cgintrinsics}} & 
{\footnotesize ClothInNet } \\
\end{tabular}
\vspace{-3mm}
\caption{ \textbf{Qualitative comparisons on real world clothing images. } 
For each method, the first row is reflectance and the second row is shading. 
Our model significantly removes texture-copying artifacts while retaining tiny details. 
The CGIntrinsics model is trained on CGI, IIW and SAW. 
}
\label{fig:qualitative4} 
\vspace{-3mm}
\end{figure*}

\clearpage
\bibliographystyle{splncs04}
\bibliography{egbib}